\documentclass[letterpaper,journal]{IEEEtran}

\usepackage{amsmath,amsfonts,amssymb}
\usepackage{algorithm}
\usepackage{algorithmic}
\usepackage{array}
\usepackage{booktabs}
\usepackage[caption=false,font=normalsize,labelfont=sf,textfont=sf]{subfig}
\usepackage{textcomp}
\usepackage{stfloats}
\usepackage{placeins}
\usepackage{url}
\usepackage{graphicx}
\usepackage{adjustbox}
\usepackage{cite}
\usepackage{bm}
\usepackage{makecell}
\usepackage{multirow}
\usepackage{xspace}
\usepackage[table,dvipsnames,xcdraw]{xcolor}
\usepackage{siunitx}
\usepackage[colorlinks=true,linkcolor=blue,citecolor=green,urlcolor=blue]{hyperref}
\usepackage{orcidlink}
\usepackage{sharpa_branding}

\SharpaBrandingOn

\makeatletter
\def\tablename{Table}
\let\orig@makecaption\@makecaption
\long\def\@makecaption#1#2{%
\ifx\@captype\@IEEEtablestring%
\footnotesize\bgroup\par\@IEEEtabletopskipstrut%
\setbox\@tempboxa\hbox{\normalfont\footnotesize {#1.}\nobreakspace #2}%
\ifdim \wd\@tempboxa >\hsize%
\setbox\@tempboxa\hbox{\normalfont\footnotesize {#1.}\nobreakspace}%
\parbox[t]{\hsize}{\normalfont\footnotesize\noindent\unhbox\@tempboxa#2}%
\else%
\hbox to\hsize{\normalfont\footnotesize\box\@tempboxa\hfil}%
\fi%
\par\addvspace{0.2\baselineskip}\egroup%
\@IEEEtablecaptionsepspace%
\else%
\orig@makecaption{#1}{#2}%
\fi}
\makeatother

\newcommand{\method}{TacBPM\xspace}
\definecolor{tacTitleColor}{HTML}{D95F02}  \definecolor{bpmTitleColor}{HTML}{1F6FB2}  
\definecolor{websiteLinkColor}{HTML}{0000FF}
\newcommand{\methodtitle}{\textcolor{tacTitleColor}{Tac}\textcolor{bpmTitleColor}{BPM}}
\newcommand{\projectwebsite}{\href{https://tacbpm.github.io}{\textcolor{websiteLinkColor}{\texttt{\textbf{https://tacbpm.github.io}}}}}
\newcommand{\latent}{\mathbf{z}}
\newcommand{\obs}{\mathbf{o}}
\newcommand{\prop}{\mathbf{x}}
\newcommand{\task}{\mathbf{g}}
\newcommand{\act}{\mathbf{a}}
\newcommand{\prior}{p_\theta}
\newcommand{\encoder}{q_\phi}
\newcommand{\decoder}{\pi_\psi}
\definecolor{colorFst}{RGB}{214, 238, 205}
\definecolor{colorSnd}{RGB}{235, 246, 218}
\definecolor{colorTrd}{RGB}{255, 248, 205}
\newcommand{\fs}{\cellcolor{colorFst}}
\newcommand{\nd}{\cellcolor{colorSnd}}
\newcommand{\rd}{\cellcolor{colorTrd}}

\begin{document}
\pagestyle{empty}
\title{\LARGE \bf
\methodtitle: A Tactile-conditioned Behavior Prior Model for Dexterous Reorientation
}

\author{Jie Yin$^{\dagger}$, Wanli Xing, Zeyuan Zhao, Xuezhou Zhu, Zhijie Deng, Kaifeng Zhang%
\thanks{$^{\dagger}$Corresponding to: Jie Yin (\href{mailto:jie.yin@sharpa.com}{jie.yin@sharpa.com}).}%
}

\IEEEaftertitletext{%
\vspace{-4em}
{\centering
  \normalsize
  Sharpa Robotics\\
  Project Website:
  \href{https://tacbpm.github.io}{\textcolor{websiteLinkColor}{\texttt{\textbf{https://tacbpm.github.io}}}}
\par}
\vspace{0.2em}
}

\maketitle
\thispagestyle{empty}
\begin{abstract}
Dexterous in-hand manipulation requires policies that coordinate high-DoF hand joints through intermittent, contact-rich interaction. Beyond target-orientation tracking, such policies must discover finger gaits that preserve object stability while adapting to geometry, anisotropy, pose, contact, and sensing changes. We propose \method, a tactile-conditioned behavior prior model for dexterous reorientation. \method distills multi-scale sphere specialists into a latent controller and lets downstream policies reuse the fixed tactile prior through residual latent actions, reducing renewed exploration from raw joint commands. The prior conditions on tactile-proprioceptive history so latent behavior reflects the current hand-object interaction. We evaluate arbitrary-pose transfer across anisotropic objects, commanded-axis rotation, and an arm-hand Grasp-to-AnyPose task in which the robot must grasp, lift, transport, and reach goal poses for novel tool geometries and generalized placements. Extensive experiments demonstrate that the proposed method accelerates training and enables stable policies where matched raw-action PPO remains near failure, with successful sim-to-real transfer in in-hand and arm-hand tasks.
\end{abstract}


\section{Introduction}
\IEEEPARstart{D}{exterous} robotic hands can rotate, stabilize, and regrasp objects without external fixtures, making them relevant to service robots, industrial manipulation, assistive systems, and teleoperation~\cite{Rajeswaran2018Dexterous,OpenAI2019Dexterous,Chen2022Reorientation}. Real deployments, however, require control policies to handle variation in shape, scale, pose, friction, and sensing quality~\cite{Qi2023RotateIt,handa2023dextreme,yang2024anyrotate}. This setting remains difficult because successful in-hand manipulation is not only a sequence of joint targets; it is a closed-loop contact strategy that must adapt to object motion, hand configuration, and fingertip feedback~\cite{Yin2023TouchDexterity,Qi2023RotateIt,Chen2026PTLD}.

In this paper, we study \emph{dexterous reorientation}: contact-rich manipulation problems in which a robot must change an object's orientation while maintaining controllable hand-object contact. We consider three progressively broader settings. \emph{In-Hand-to-AnyPose} asks the hand to reach arbitrary target orientations from an established in-hand state, where anisotropic objects introduce pose-dependent contact modes through edges, elongated bodies, and irregular geometry. \emph{Axis-Conditioned Rotation} replaces target orientations with compact signed-axis commands, requiring one policy to realize six rotation modes while preserving the grasp and maintaining stable contact. \emph{Grasp-to-AnyPose} extends the problem to an arm-hand setting that couples grasp acquisition, transport, and goal-pose reaching. Together, these settings move beyond single-axis spinning or fixed-grasp cube rotation~\cite{handa2023dextreme,yang2024anyrotate}: actions that roll a sphere may wedge a block, slip on a compliant object, or destabilize a transported grasp. They therefore test arbitrary-pose intent, transfer across scale and anisotropic shape, and validation across in-hand and arm-hand embodiments.

Large-scale reinforcement learning has produced capable dexterous manipulation policies in simulation~\cite{Schulman2017-ft,Makoviychuk2021-sx}. With appropriate rewards, domain randomization, and privileged object information, specialist policies can reorient objects to arbitrary poses~\cite{OpenAI2019Dexterous,Chen2022Reorientation,handa2023dextreme}. The limitation is reusability: many systems remain strongest in narrower regimes, such as a fixed object shape, a known grasp state, a single rotation style, visual or privileged object-pose feedback, or direct transfer of a complete pretrained actor~\cite{handa2023dextreme,bhardwaj2026viserdex,Kedia2026SimToolReal,Lum2026Play2Perfect}. Downstream learning must then re-explore low-level joint commands, where random actions often break contact or fail to discover finger gaits that stabilize the object while inducing rotation.

A reusable behavior prior can reduce this burden by compressing contact-rich motion patterns into latent actions~\cite{Peng2022-vr,Tessler_undated-zi,Luo2024Pulse}. Prior work in physics-based character control shows that broad humanoid motions can be distilled into latent motor priors~\cite{Luo2024Pulse}, but dexterous reorientation poses a different fine-manipulation problem. The prior cannot be only a proprioceptive motion manifold: it must react to intermittent fingertip contact, load transfer, contact location, slip, and scale-induced contact regimes~\cite{Yin2023TouchDexterity,Qi2023RotateIt,Chen2026PTLD}. An effective hand prior should therefore change when a fingertip gains or loses contact, force shifts across fingers, or the object begins to slip.

We propose \method, a tactile-conditioned behavior prior model for dexterous in-hand reorientation. For the in-hand benchmarks, \method distills eight sphere specialists spanning scale factors from $0.3$ to $1.0$ into a compact variational latent controller, exposing the hand to different apertures, contact-force patterns, and rolling and regrasping regimes while keeping teacher geometry simple. For arm-hand validation, we leverage the same method paradigm with scale-randomized rubber-hammer teachers and evaluate downstream policies on diverse tool geometries. Unlike a purely proprioceptive prior, \method conditions its latent behavior on recent hand state, previous targets, tactile contact magnitudes and positions.

Downstream policies keep the tactile prior frozen and learn residual latent corrections from task observations. In the in-hand setting, the latent-to-action mapping is finetuned with the high-level policy to adapt to new geometries without changing the contact-conditioned prior. This separates reusable contact behavior from task intent: the prior anchors exploration near contact-preserving hand-object interaction patterns, while the downstream policy selects low-dimensional corrections. As a result, \method bootstraps learning where matched raw-action PPO rarely reaches contact-stable behavior, and yields substantially higher success on complex reorientation tasks within the same budget.
We highlight our contributions as:
\begin{itemize}
    \item We introduce a tactile-conditioned behavior prior for in-hand dexterous reorientation, distilling multi-scale sphere specialists into a variational prior-decoder controller that adapts latent behavior to contact magnitude, contact position, and scale-induced contact changes.
    \item We reuse the learned prior within a residual latent control framework for downstream reinforcement learning, keeping the prior frozen while allowing decoder adaptation for new in-hand geometries.
    \item We evaluate unseen sphere-scale extrapolation, anisotropic-object transfer, commanded-axis rotation, and arm-hand goal-pose reaching, showing improved learning performance and successful sim-to-real transfer.
\end{itemize}

\section{Related Work}
\vspace{-1mm}
\subsection{Reusable Skill Representations}
\vspace{-1mm}

Reusable latent action spaces provide structured exploration for high-dimensional control, especially in physics-based character and humanoid control~\cite{Peng2022-vr,Tessler_undated-zi,Merel2018-bq,Yao2022-as,Won2022-jy,Zhu2023-nn,Luo2024Pulse}. These priors mainly model whole-body locomotion from proprioceptive states and do not explicitly condition reusable skills on fingertip contact or hand-object interaction. In contrast, \method distills dexterous sphere specialists into a tactile-conditioned prior, keeps that prior fixed downstream, and adapts the decoder for new geometries. Its specialist-to-generalist online distillation aggregates multiple sphere teachers into a reusable skill latent, then exposes residual latent corrections rather than transferring a complete executable actor.

\vspace{-3mm}
\subsection{Tactile-Proprioceptive Control}
\vspace{-1mm}

Tactile sensing provides direct evidence of contact state, geometry distribution, and incipient slip. Touch-only, visuotactile, and tactile-representation systems use contact signals for object rotation, reorientation, translation, fine-grained manipulation, and state inference under occlusion~\cite{Yin2023TouchDexterity,Qi2023RotateIt,Yuan2024RobotSynesthesia,yin2025learning,Guzey2023TDex,suresh2024neuralfeels,Huang2024ViTac}. PTLD deploys privileged-sensor policies in an instrumented real-world setup with external object tracking, then distills privileged latents into tactile state estimators for rotation and reorientation~\cite{Chen2026PTLD}; this avoids dense tactile simulation, but focuses on recovering privileged policy state for deployment rather than learning a reusable tactile behavior prior for downstream residual-latent control, and it does not study decoder finetuning as a transfer mechanism. WM-Craftnet instead learns a visuotactile world model as recurrent task context for robust in-hand manipulation~\cite{Yin2026WMCraftnet}. These works establish tactile sensing as an important observation source, whereas \method uses tactile and proprioceptive history to parameterize latent control itself: the prior and decoder change the latent action distribution with current contact magnitude, contact position, previous targets, and joint state.

\vspace{-4mm}
\subsection{Dexterous In-Hand Rotation and Reorientation}
\vspace{-1mm}

Dexterous in-hand rotation and reorientation have been advanced by large-scale reinforcement learning, domain randomization, rapid motor adaptation, visual perception transfer, touch-aware policies, learned dynamics adaptation, and simple-to-complex curricula~\cite{Rajeswaran2018Dexterous,OpenAI2019Dexterous,Chen2022Reorientation,handa2023dextreme,Qi2022Hora,Qi2023RotateIt,bhardwaj2026viserdex,yang2024anyrotate,liu2025dexndm,wang2024lessons,qi2025simple,wan2025dexremoe}. Much of this literature optimizes task-specific policies in high-DoF raw-action spaces, so changing object shapes, sensing assumptions, or task definitions often requires retraining and reward retuning. Representative systems address cube reorientation, tactile multi-axis rotation, visual reorientation, or simple-to-complex curricula~\cite{handa2023dextreme,yang2024anyrotate,bhardwaj2026viserdex,qi2025simple}. Recent arm-hand systems formulate tool use and assembly as goal-pose reaching~\cite{Kedia2026SimToolReal,Lum2026Play2Perfect}. \method is complementary: it compresses contact-rich motion patterns into a tactile-conditioned behavior prior for residual-latent reuse.

\section{Method}
\subsection{Overview}
\method compresses contact-rich finger gaits and object-interaction patterns into structured latent control. We learn a deterministic decoder $\hat{\act}=\decoder(\prop,\latent)$ and a tactile-conditioned Gaussian prior $\prior(\latent|\prop)$, then let downstream policies act through residual latent codes instead of directly in the 22-D action space. The in-hand setting trains multi-scale sphere specialists, distills their behavior into an encoder-prior-decoder controller, and reuses the fixed prior on downstream tasks while adapting the decoder. The arm-hand study instantiates the same teacher-distillation and residual-latent paradigm with scale-randomized rubber-hammer teachers.

\begin{figure*}[t]
    \centering
    \begin{adjustbox}{width=0.99\linewidth,center}
        \includegraphics{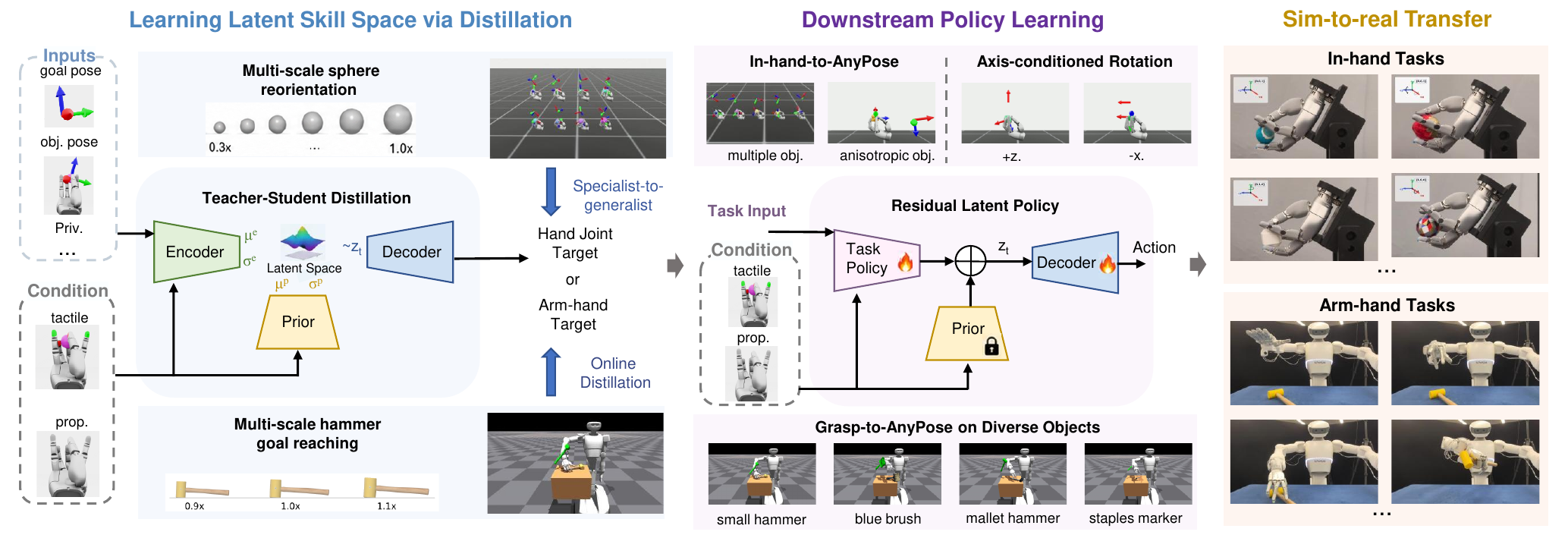}
    \end{adjustbox}
    \caption{\textbf{Overview of \method.} The pipeline distills specialist behavior into a tactile-conditioned latent controller, then applies a prior-centered residual-latent interface to the corresponding downstream tasks.}
    \label{fig:pipeline}
    \vspace{-3mm}
\end{figure*}

\vspace{-2mm}
\subsection{Specialist Policy Training}
\vspace{-1mm}

The in-hand specialist stage provides controlled contact-behavior coverage. We train sphere reorientation specialists with proximal policy optimization~\cite{Schulman2017-ft} in Isaac Sim using Isaac Lab's massively parallel robot-learning infrastructure~\cite{Mittal2025IsaacLab}. The nominal sphere diameter is \SI{8}{cm}, and the eight specialists use scale factors $\{0.3,0.4,\ldots,1.0\}$. Each specialist observes a teacher-compatible state $\obs_t$ containing tactile-proprioceptive history, task goal information, and teacher-only privileged object information. Varying scale changes hand aperture, contact distribution, and rolling and regrasping regimes while keeping object geometry simple; using spheres isolates contact-regime diversity from shape complexity before the prior is reused on more challenging in-hand reorientation tasks.

Online multi-teacher distillation aligns supervision with the student's rollout distribution. Environments are assigned to the eight sphere-scale teachers in a strided batch layout. At each step, the assigned teacher produces the corresponding low-level target action, and all targets are interleaved into a single training batch. This DAgger-style procedure preserves scale-specific expert behavior while avoiding an offline action dataset that may not match the student's induced states.

\vspace{-2mm}
\subsection{Tactile-Conditioned State Representation}
\vspace{-1mm}

The student state $\prop_t$ is a compact tactile-proprioceptive history. Each frame contains normalized hand joint positions $\mathbf{q}$, previous control targets $\mathbf{u}$, five smoothed tactile contact magnitudes $\mathbf{c}^{f}$, and five 3D tactile contact positions $\mathbf{c}^{p}$. We stack the most recent three frames into a 192-dimensional state:
\begin{equation}
\begin{aligned}
\prop_t=\big[
&\mathbf{q}_{t-2:t},\mathbf{u}_{t-2:t},\\
&\mathbf{c}^{f}_{t-2:t},\mathbf{c}^{p}_{t-2:t}
\big].
\end{aligned}
\end{equation}
For In-Hand-to-AnyPose, the task observation $\task_t$ is the relative target orientation.

In the main MLP-concatenation setting, $\mathbf{h}_t=f_\eta(\prop_t)$ denotes the normalized tactile-proprioceptive feature formed from this history. During distillation, the encoder additionally receives an object longest-side cue, whereas the deployed prior and decoder use $\mathbf{h}_t$; scale variation is therefore learned through teacher assignment, contact state, and latent supervision. Consequently, the same task goal can induce different latent codes under stable pinch, weak contact, or impending contact loss.

\vspace{-2mm}
\subsection{Specialist-to-Generalist Variational Distillation}
\vspace{-1mm}

\method distills specialist actions into an encoder-prior-decoder latent controller. Let $\mathbf{h}_t$ denote the tactile-proprioceptive feature extracted from $\prop_t$. The encoder predicts a task-conditioned posterior, the prior predicts a task-agnostic latent distribution without the goal, and the decoder maps the latent-state pair to a deterministic hand position target:
\begin{equation}
\begin{aligned}
\left(\bm{\mu}^{\mathrm{enc}}_t,\bm{\sigma}^{\mathrm{enc}}_t\right)&=\encoder(\mathbf{h}_t,\task_t),\\
\left(\bm{\mu}^{\mathrm{prior}}_t,\bm{\sigma}^{\mathrm{prior}}_t\right)&=\prior(\mathbf{h}_t),\\
\hat{\act}_t&=\decoder(\mathbf{h}_t,\latent_t).
\end{aligned}
\end{equation}
The posterior and prior are diagonal Gaussians. During distillation, $\latent_t$ is sampled from the posterior with the reparameterization trick; deterministic evaluation uses the corresponding mean:
\begin{equation}
\begin{aligned}
\encoder(\latent_t|\prop_t,\task_t)
&=\mathcal{N}(\bm{\mu}^{\mathrm{enc}}_t,\operatorname{diag}((\bm{\sigma}^{\mathrm{enc}}_t)^2)),\\
\prior(\latent_t|\prop_t)
&=\mathcal{N}(\bm{\mu}^{\mathrm{prior}}_t,\operatorname{diag}((\bm{\sigma}^{\mathrm{prior}}_t)^2)).
\end{aligned}
\end{equation}
During online distillation, the student rolls out in the environment and the assigned sphere-scale specialist provides $\act^{\star}_t$ at the same state. The objective combines action matching, KL regularization to the learned prior, and temporal smoothness over consecutive posterior means:
\begin{equation}
\begin{aligned}
\mathcal{L} =
&\|\hat{\act}_t-\act^{\star}_t\|_2^2
+\beta\,D_{\mathrm{KL}}\!\left(\encoder(\latent_t|\prop_t,\task_t)\,\|\,\prior(\latent_t|\prop_t)\right) \\
&+\lambda\|\bm{\mu}^{\mathrm{enc}}_t-\bm{\mu}^{\mathrm{enc}}_{t-1}\|_2^2 .
\end{aligned}
\end{equation}
The KL term trains the prior to predict task-relevant latent codes without the task goal, and the temporal term discourages discontinuous latent jumps.

\vspace{-2mm}
\subsection{Residual Latent Downstream Control}
\vspace{-1mm}

Downstream learning uses the distilled prior as a fixed action-space anchor. The tactile state normalizer and prior network are frozen. A PPO policy receives the normalized tactile-proprioceptive state and task observation, samples a residual latent action $\Delta\latent_t$ from its Gaussian policy, and executes
\begin{equation}
\begin{aligned}
\latent^{\mathrm{task}}_t&=\bm{\mu}^{\mathrm{prior}}_t+\Delta\latent_t,\\
\act^{\mathrm{task}}_t&=\decoder(\mathbf{h}_t,\latent^{\mathrm{task}}_t).
\end{aligned}
\end{equation}
This residual formulation constrains exploration near contact-rich patterns from the tactile-conditioned prior while still allowing task-driven deviations. In the in-hand downstream experiments, only the downstream policy and value heads and the VAE decoder network and output head are trainable; the prior and state normalizers remain fixed. Decoder finetuning adapts latent-command realization to new geometry without changing the contact-conditioned prior, and a frozen-decoder ablation isolates this effect.

\vspace{-2mm}
\subsection{Arm-Hand Embodiment Validation}
\label{sec:arm_hand_method}
\vspace{-1mm}

We further validate the same prior-centered training paradigm on an arm-hand embodiment with a right arm and right hand, which we call Grasp-to-AnyPose. The robot must grasp an object, transport it, and reorient it to a sampled goal pose, coupling arm motion with contact-stable dexterous hand behavior. Because this embodiment changes the action space and observations, we train rubber-hammer teachers with continuous scale randomization from $0.9$ to $1.1$, distill a separate tactile-conditioned latent controller, and evaluate downstream generalization to small hammer, blue brush, staples marker, and mallet hammer. This experiment tests transfer of the paradigm rather than reuse of the in-hand sphere prior. The task varies object category and scale, initial placement, goal workspace, goal-to-goal translations, and target orientation. The prior is conditioned on arm-hand proprioception, previous targets, palm pose, object-relative keypoints, and five-fingertip contact signals; the downstream policy selects residual latent commands for unseen tool geometries.

Inspired by the staged hand-arm rewards in DexPBT and the goal-pose shaping in SimToolReal~\cite{Petrenko2023DexPBT,Kedia2026SimToolReal}, the Grasp-to-AnyPose reward first encourages fingertip approach and lifting, then rewards reductions in the maximum object-to-goal keypoint distance and adds a sparse goal bonus when all keypoints fall within tolerance:
\begin{equation}
\begin{aligned}
d_{\mathrm{kp}} &= \max_i \left\|\mathbf{k}^{\mathrm{obj}}_{i,t}-\mathbf{k}^{\mathrm{goal}}_i\right\|_2,\\
r_t &= r_{\mathrm{approach}}+r_{\mathrm{lift}}+r_{\mathrm{kp}}+\mathbb{I}[d_{\mathrm{kp}}<\epsilon_{\mathrm{goal}}]r_{\mathrm{bonus}}-r_{\mathrm{reg}} .
\end{aligned}
\end{equation}
Penalties on arm and hand actions, object velocity, and arm-table collisions regularize the behavior. Downstream training uses 8192 environments, success resets after each reached goal, a sparse goal bonus, and no object-angular-velocity penalty.

\vspace{-2mm}
\subsection{Training Details}
\label{sec:implementation_details}
\vspace{-1mm}

The in-hand controller uses 22-DoF position-target actions, a 16-D Gaussian latent action, and encoder, prior, and decoder MLPs with hidden sizes $(512,256,128)$. The tactile-proprioceptive state is a three-frame history of joints, previous targets, contact magnitudes and contact locations. Distillation uses online teacher action matching with KL annealing and temporal smoothing, so the latent controller reconstructs contact-stable expert behavior while keeping the prior regularized. Unless stated otherwise, in-hand distillation and downstream PPO use 8192 environments, Adam with learning rate $3\times10^{-4}$, horizon length 8, minibatches of 32768 samples, and five PPO epochs. Axis evaluation uses six signed commands per object. The arm-hand validation uses latent-action RL-Games PPO, fingertip contacts, and a 29-D action space $\act^{\mathrm{arm\text{-}hand}}_t=[\act^{\mathrm{hand}}_t,\act^{\mathrm{arm}}_t]\in\mathbb{R}^{22+7}$.

\section{Experiments}
\label{sec:experiments}
To evaluate whether distilled contact behavior forms a reusable prior rather than a sphere-specific controller, we organize the experiments around four questions:
\begin{itemize}
    \setlength{\itemsep}{-1mm}
    \item \textbf{Q1. Prior coverage and extrapolation:} Do multi-scale sphere teachers provide contact-behavior support that transfers to unseen sphere scales and non-spherical shapes?
    \item \textbf{Q2. Complex-object in-hand transfer:} Does tactile residual-latent control with decoder adaptation improve arbitrary-pose reorientation on anisotropic objects?
    \item \textbf{Q3. Commanded-axis rotation and hardware transfer:} Can one latent controller follow compact signed-axis commands in simulation and on a real dexterous hand?
    \item \textbf{Q4. Arm-hand embodiment validation:} Can the same training paradigm support grasp acquisition, transport, and goal-pose reaching?
\end{itemize}

\vspace{-4mm}
\subsection{Experimental Setup}
\paragraph{Common simulation protocol.}
All in-hand simulation and hardware experiments use the 22-DoF SharpaWave dexterous hand, and the arm-hand experiments use the Sharpa North robot equipped with the same hand. All simulation comparisons use the same dexterous hand model, initialization protocol, and termination rules within each benchmark. The in-hand prior is trained from eight sphere specialists with scale factors $\{0.3,0.4,\ldots,1.0\}$ of a nominal \SI{8}{cm} sphere; for these in-hand evaluations, the extrapolation sphere scales ($1.1$ and $1.2$) and all additional shapes are excluded from teacher training. Unless otherwise stated, each row is evaluated over 1000 randomized attempts.

\paragraph{Common downstream control.}
Downstream policies using \method predict residual latent codes rather than low-level joint actions. The residual is added to the frozen prior mean and decoded into a 22-DoF hand command. The main in-hand setting keeps the prior and state normalizers fixed while finetuning the decoder and output head; the strict-reuse ablation freezes both prior and decoder. These comparisons separate the latent prior, tactile conditioning, and decoder adaptation effects.

\paragraph{Baselines.}
RL from scratch learns directly in the environment action space. To enable fair comparison, baselines and \method use the same reward, curriculum, environment count, PPO hyperparameters, training budget, initialization, and evaluation protocol within each benchmark. The tactile-free ablation keeps the residual-latent prior-decoder pipeline but zeros tactile force and contact-position channels before normalization. Additional complex-object baselines isolate actor-level transfer, action-space residual learning, single-scale teacher coverage, decoder finetuning, and frozen-decoder reuse.

\paragraph{Metrics and uncertainty.}
Following prior dexterous reorientation and in-hand rotation work~\cite{OpenAI2019Dexterous,Chen2022Reorientation,Qi2022Hora,Qi2023RotateIt}, we report success, drop rate, episode length, and rotation progress. A trial succeeds if the object reaches within $0.2$ rad of the target orientation before timeout or drop. Object drop is defined by a height-based reset. Steps$^\dagger$ counts failed attempts at the episode cap. Unless stated otherwise, tabulated uncertainties are 95\% confidence intervals over evaluation attempts and episodes.

\vspace{-2mm}
\subsection{Sphere-Scale In-Hand-to-AnyPose Extrapolation}
\vspace{-2mm}

We first test whether the distilled prior transfers beyond its teacher scales: whether multi-scale sphere teachers provide broad contact-behavior coverage and whether the resulting controller extrapolates to unseen scales and shapes.

\begin{figure}[t]
    \centering
    \setlength{\tabcolsep}{2pt}
    \begin{tabular}{@{}cc@{}}
    \includegraphics[width=0.48\columnwidth]{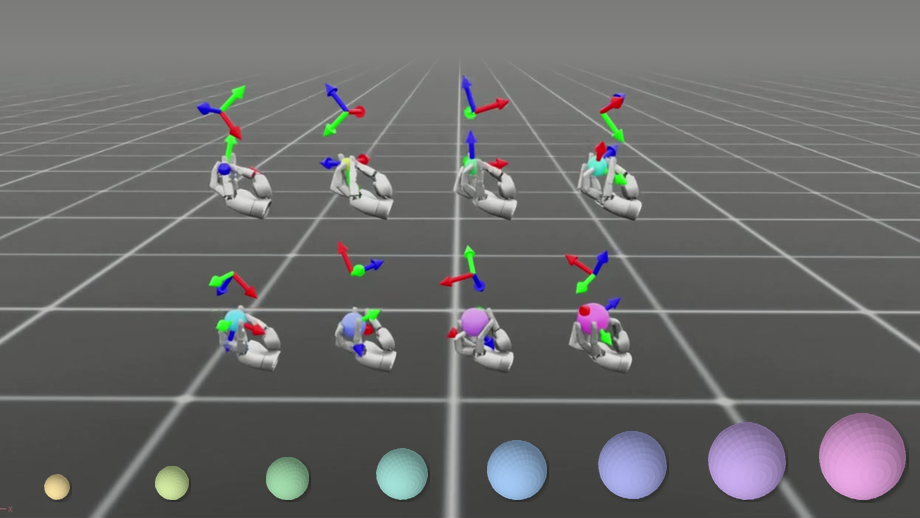} &
    \includegraphics[width=0.48\columnwidth]{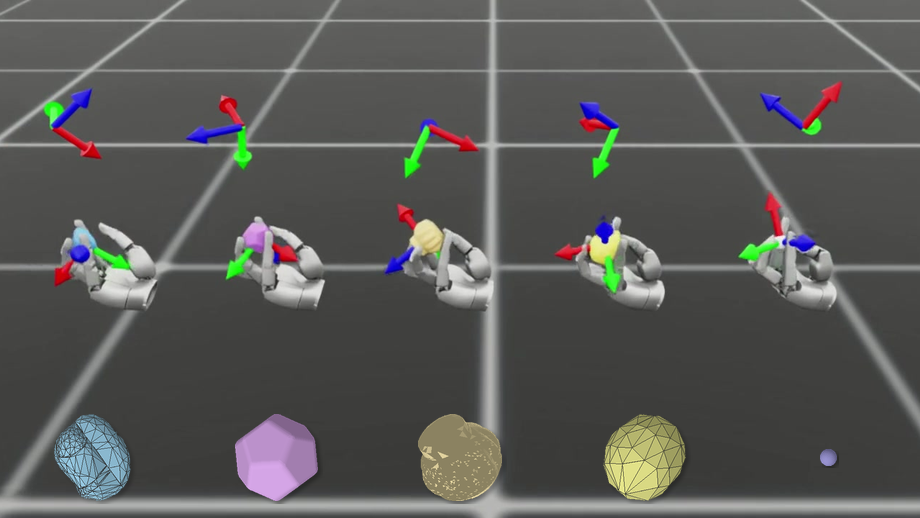} \\
    {\scriptsize (a) Multi-scale sphere teachers} &
    {\scriptsize (b) one policy for multi-object reorientation}
    \end{tabular}
    \caption{\textbf{Prior coverage and downstream reuse.} The panels show sphere-teacher rollouts and a downstream arbitrary-pose rollout on multiple object instances.}
    \label{fig:prior_coverage_overview}
    \vspace{-2mm}
\end{figure}

\noindent\textbf{Qualitative and latent evidence.}
Figure~\ref{fig:prior_coverage_overview} gives a qualitative view of the teacher-to-downstream transition. The sphere-teacher rollouts span object sizes and target orientations, requiring changes in hand aperture, fingertip placement, and rolling and regrasping strategies. The downstream rollout applies the same prior-centered interface to non-spherical objects whose contact geometry differs from the teacher family. The bottom overlays make this contrast explicit: panel (a) shows the eight teacher sphere scales, whereas panel (b) shows downstream objects with different scales, curvatures, and faceted structures. This motivates the quantitative question in Table~\ref{tab:sphere_generalist_by_scale}: the prior should preserve stable sphere manipulation while leaving enough residual freedom for new shapes.

\begin{figure}[t]
    \centering
    \begin{adjustbox}{width=0.75\columnwidth,center}
        \includegraphics{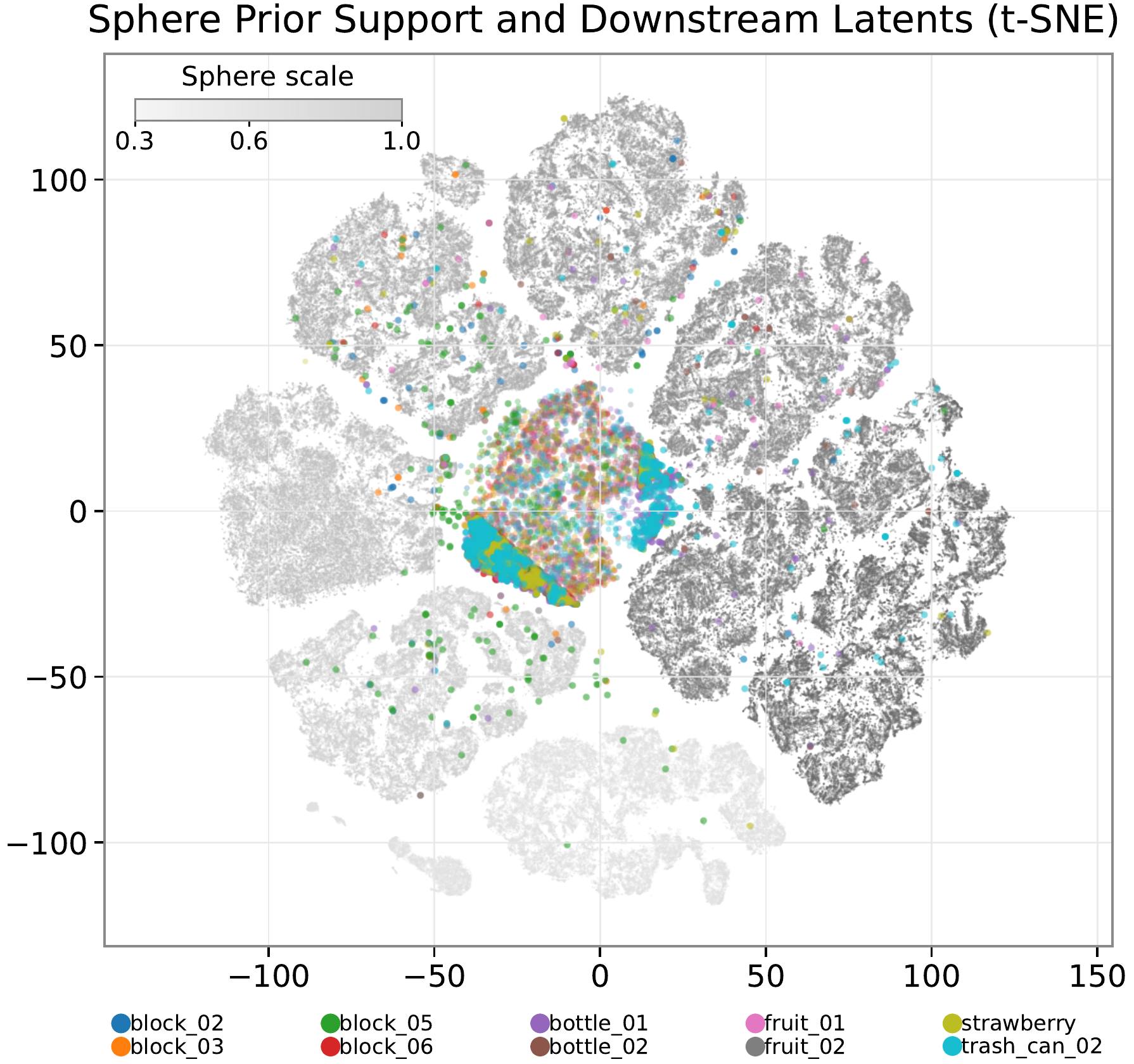}
    \end{adjustbox}
    \caption{\textbf{Qualitative latent visualization.} The t-SNE plot shows sphere-prior latent samples and executed downstream task latents.}
    \label{fig:prior_latent_tsne}
    \vspace{-2mm}

\end{figure}

Figure~\ref{fig:prior_latent_tsne} complements the rollout view with a qualitative two-dimensional projection of the 16-D latent space. Gray points are sphere-prior latent samples, and colored points are executed downstream task latents $\latent^{\mathrm{task}}_t=\bm{\mu}^{\mathrm{prior}}_t+\Delta\latent_t$. The t-SNE plot shows that many downstream commands remain near the projected sphere-prior cloud, while boundary excursions mark task-driven deviations. 

\noindent\textbf{Quantitative extrapolation.}
Table~\ref{tab:sphere_generalist_by_scale} evaluates whether the qualitative coverage translates into policy performance across seen sphere scales, unseen sphere scales, and novel shapes.
Results indicate that raw-action PPO succeeds on the easiest mid-size sphere but is unstable across most other scales under the matched protocol, suggesting that direct exploration in the 22-D action space is a bottleneck in this setting. The learned prior-decoder framework without tactile input preserves much of the teacher family on seen scales, increasing average success from $26.03\%$ to $93.67\%$. Tactile conditioning adds only a small gain on seen scales ($93.67\%$ to $95.62\%$), matching the visual observation that these states remain close to the teacher support.
\begin{table}[t]
    \caption{\textbf{Sphere-scale and unseen-object evaluation of the generalist prior ablations.} Rows include seen sphere scales, unseen sphere scales, and object shapes excluded from teacher training.}
\label{tab:sphere_generalist_by_scale}
\centering
{%
\setlength{\tabcolsep}{2.4pt}
\renewcommand{\arraystretch}{1.05}
\resizebox{\columnwidth}{!}{%
\begin{tabular}{lcccccc}
\toprule
\multirow{2}{*}{Case} & \multicolumn{2}{c}{From scratch} & \multicolumn{2}{c}{No tactile} & \multicolumn{2}{c}{Ours} \\
\cmidrule(lr){2-3}\cmidrule(lr){4-5}\cmidrule(lr){6-7}
& Succ. $\uparrow$ & Steps$^\dagger$ $\downarrow$ & Succ. $\uparrow$ & Steps$^\dagger$ $\downarrow$ & Succ. $\uparrow$ & Steps$^\dagger$ $\downarrow$ \\
\midrule
sphere ($0.3$) & $24.50{\pm}2.67\%$ & $79.40{\pm}2.27$ & $92.90{\pm}1.59\%$ & $35.03{\pm}1.59$ & {\fs\bfseries\boldmath $97.30{\pm}1.01\%$} & {\fs\bfseries\boldmath $31.07{\pm}1.26$} \\
sphere ($0.4$) & $25.20{\pm}2.69\%$ & $79.81{\pm}2.21$ & $96.10{\pm}1.20\%$ & $29.76{\pm}1.16$ & {\fs\bfseries\boldmath $98.40{\pm}0.78\%$} & {\fs\bfseries\boldmath $27.58{\pm}0.94$} \\
sphere ($0.5$) & $99.80{\pm}0.28\%$ & {\fs\bfseries\boldmath $23.28{\pm}0.30$} & {\fs\bfseries\boldmath $100.00{\pm}0.00\%$} & $23.42{\pm}0.18$ & $99.90{\pm}0.20\%$ & $23.45{\pm}0.27$ \\
sphere ($0.6$) & $20.40{\pm}2.50\%$ & $84.18{\pm}2.00$ & $96.50{\pm}1.14\%$ & $33.33{\pm}1.26$ & {\fs\bfseries\boldmath $98.90{\pm}0.65\%$} & {\fs\bfseries\boldmath $29.72{\pm}0.98$} \\
sphere ($0.7$) & $14.60{\pm}2.19\%$ & $89.27{\pm}1.67$ & $94.20{\pm}1.45\%$ & $38.31{\pm}1.43$ & {\fs\bfseries\boldmath $97.00{\pm}1.06\%$} & {\fs\bfseries\boldmath $35.29{\pm}1.11$} \\
sphere ($0.8$) & $10.20{\pm}1.88\%$ & $92.95{\pm}1.47$ & $92.90{\pm}1.59\%$ & $41.73{\pm}1.36$ & {\fs\bfseries\boldmath $96.80{\pm}1.09\%$} & {\fs\bfseries\boldmath $39.34{\pm}1.17$} \\
sphere ($0.9$) & $8.40{\pm}1.72\%$ & $93.95{\pm}1.30$ & $91.40{\pm}1.74\%$ & $50.63{\pm}1.50$ & {\fs\bfseries\boldmath $92.00{\pm}1.68\%$} & {\fs\bfseries\boldmath $49.81{\pm}1.50$} \\
sphere ($1.0$) & $5.10{\pm}1.36\%$ & $96.11{\pm}1.05$ & {\fs\bfseries\boldmath $85.40{\pm}2.19\%$} & $61.13{\pm}1.66$ & $84.70{\pm}2.23\%$ & {\fs\bfseries\boldmath $60.88{\pm}1.66$} \\
\textbf{Seen Avg.} & $26.03{\pm}21.29\%$ & $77.37{\pm}21.21$ & $93.67{\pm}2.97\%$ & $36.67{\pm}11.68$ & {\fs\bfseries\boldmath $95.62{\pm}3.47\%$} & {\fs\bfseries\boldmath $34.64{\pm}11.72$} \\
\midrule
sphere ($1.1$) & $4.90{\pm}1.34\%$ & $96.20{\pm}1.05$ & $55.50{\pm}3.08\%$ & $85.10{\pm}1.39$ & {\fs\bfseries\boldmath $62.00{\pm}3.01\%$} & {\fs\bfseries\boldmath $83.10{\pm}1.49$} \\
sphere ($1.2$) & $3.30{\pm}1.11\%$ & {\fs\bfseries\boldmath $97.52{\pm}0.84$} & $1.70{\pm}0.80\%$ & $100.04{\pm}0.44$ & {\fs\bfseries\boldmath $13.30{\pm}2.11\%$} & $99.80{\pm}0.98$ \\
block\_01 & $18.30{\pm}2.40\%$ & $85.26{\pm}1.95$ & $58.20{\pm}3.06\%$ & $72.56{\pm}2.07$ & {\fs\bfseries\boldmath $79.50{\pm}2.50\%$} & {\fs\bfseries\boldmath $61.10{\pm}1.92$} \\
trash\_can\_01 & $11.70{\pm}1.99\%$ & $90.62{\pm}1.61$ & $14.60{\pm}2.19\%$ & $94.56{\pm}1.25$ & {\fs\bfseries\boldmath $79.52{\pm}1.05\%$} & {\fs\bfseries\boldmath $61.07{\pm}0.90$} \\
switch\_01 & $13.80{\pm}2.14\%$ & $89.30{\pm}1.68$ & $28.70{\pm}2.81\%$ & $85.27{\pm}1.72$ & {\fs\bfseries\boldmath $47.10{\pm}3.10\%$} & {\fs\bfseries\boldmath $80.49{\pm}1.90$} \\
box\_01 & $13.50{\pm}2.12\%$ & $89.69{\pm}1.65$ & $29.00{\pm}2.81\%$ & $90.24{\pm}1.68$ & {\fs\bfseries\boldmath $41.80{\pm}3.06\%$} & {\fs\bfseries\boldmath $88.24{\pm}1.99$} \\
bottle\_03 & $11.80{\pm}2.00\%$ & $90.55{\pm}1.62$ & $57.70{\pm}3.06\%$ & $69.21{\pm}1.91$ & {\fs\bfseries\boldmath $77.80{\pm}2.58\%$} & {\fs\bfseries\boldmath $59.99{\pm}1.79$} \\
strawberry & $14.50{\pm}2.18\%$ & $88.40{\pm}1.76$ & $75.00{\pm}2.69\%$ & $52.26{\pm}1.87$ & {\fs\bfseries\boldmath $82.00{\pm}2.38\%$} & {\fs\bfseries\boldmath $47.83{\pm}1.69$} \\
\textbf{Unseen Avg.} & $11.47{\pm}3.47\%$ & $90.94{\pm}2.80$ & $40.05{\pm}17.51\%$ & $81.16{\pm}10.81$ & {\fs\bfseries\boldmath $60.38{\pm}17.06\%$} & {\fs\bfseries\boldmath $72.71{\pm}12.27$} \\
\bottomrule
\end{tabular}}%
}
\end{table}

The advantage of tactile conditioning becomes clearer under contact-regime shift. On unseen scales and additional shapes, \method increases average success from $40.05\%$ to $60.38\%$ and reduces capped steps from $81.16$ to $72.71$. The no-tactile baseline is particularly weak on sphere ($1.2$) and trash\_can\_01, where success falls to $1.70\%$ and $14.60\%$, suggesting that proprioception alone does not identify contact regimes that depart from the teacher support. Together with the qualitative view in Fig.~\ref{fig:prior_latent_tsne}, these results suggest that residual policies benefit from a contact-stable latent anchor while tactile inputs select task-appropriate deviations. The remaining gap on sphere ($1.2$) and several novel shapes indicates the limit of extrapolating from a frozen sphere-trained prior.

\vspace{-2mm}
\subsection{In-Hand-to-AnyPose Transfer on Complex Objects}
\vspace{-2mm}

\subsubsection{Task and baselines}
We next evaluate downstream residual policies on anisotropic objects whose contact geometry differs from the sphere teachers. This In-Hand-to-AnyPose setting is difficult because broad target orientations interact with object edges, elongated bodies, irregular shapes, and pose-dependent contact affordances. The reward combines dense rotational progress, a goal-tolerance bonus, and penalties on position drift, large actions, non-nominal hand posture, torque, work, and falls. All methods use the matched protocol described above.

\begin{figure}[t]
    \centering
    \setlength{\tabcolsep}{1.2pt}
    \begin{tabular}{@{}ccc@{}}
    \includegraphics[width=0.318\columnwidth]{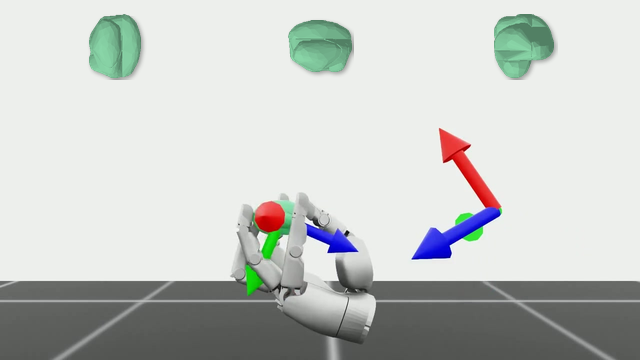} &
    \includegraphics[width=0.318\columnwidth]{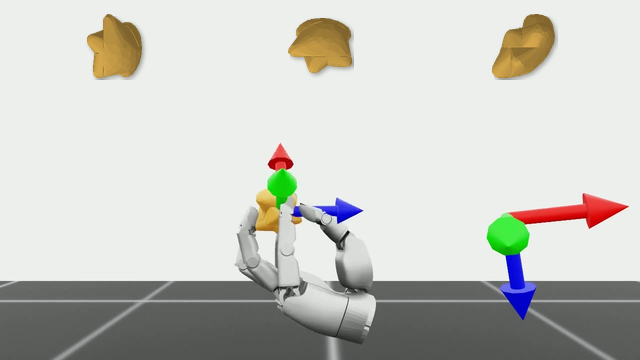} &
    \includegraphics[width=0.318\columnwidth]{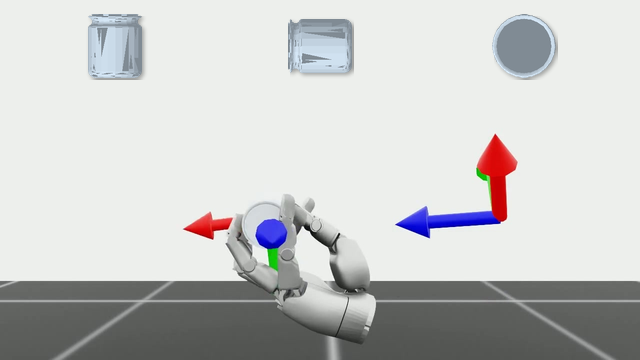} \\
    {\scriptsize (a) block\_03} & {\scriptsize (b) block\_05} & {\scriptsize (c) bottle\_01} \\
    \includegraphics[width=0.318\columnwidth]{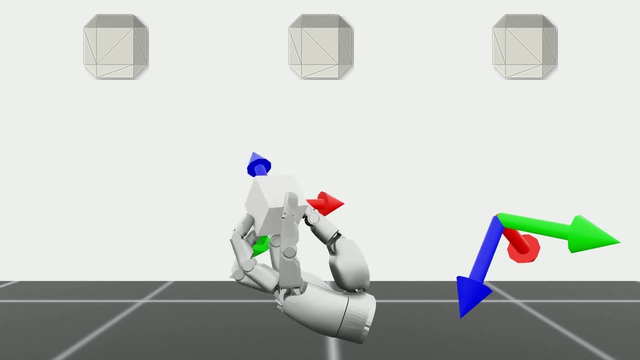} &
    \includegraphics[width=0.318\columnwidth]{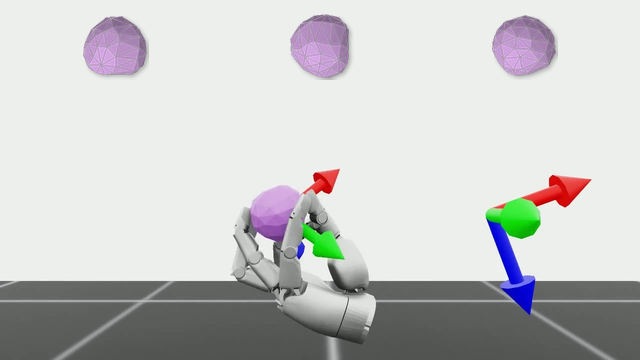} &
    \includegraphics[width=0.318\columnwidth]{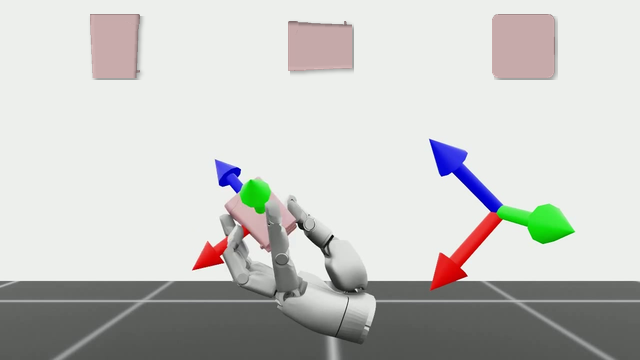} \\
    {\scriptsize (d) corner block} & {\scriptsize (e) fruit\_02} & {\scriptsize (f) trash\_can\_02}
    \end{tabular}
    \caption{\textbf{Qualitative examples from arbitrary-pose reorientation on complex objects.} The panels show representative downstream rollouts across six object geometries.}
    \label{fig:downstream_objects_qualitative}
    \vspace{-2mm}

\end{figure}

\begin{table*}[t]
    \caption{\textbf{Downstream reorientation on complex objects.} The table reports success and episode length for transfer baselines and \method variants.}
    \label{tab:downstream_objects}
    \centering
    \resizebox{\textwidth}{!}{%
    \begin{tabular}{lcccccccccccc}
    \toprule
    & \multicolumn{2}{c}{block\_03} & \multicolumn{2}{c}{block\_05} & \multicolumn{2}{c}{bottle\_01} & \multicolumn{2}{c}{fruit\_02} & \multicolumn{2}{c}{trash\_can\_02} & \multicolumn{2}{c}{Multi} \\
    \cmidrule(lr){2-3}\cmidrule(lr){4-5}\cmidrule(lr){6-7}\cmidrule(lr){8-9}\cmidrule(lr){10-11}\cmidrule(lr){12-13}
    Method & Succ. $\uparrow$ & Steps $\downarrow$ & Succ. $\uparrow$ & Steps $\downarrow$ & Succ. $\uparrow$ & Steps $\downarrow$ & Succ. $\uparrow$ & Steps $\downarrow$ & Succ. $\uparrow$ & Steps $\downarrow$ & Succ. $\uparrow$ & Steps $\downarrow$ \\
    \midrule
    RL from scratch & $1.40{\pm}0.73\%$ & $98.73{\pm}0.66$ & $2.10{\pm}0.89\%$ & $98.09{\pm}0.81$ & $1.70{\pm}0.80\%$ & $98.42{\pm}0.74$ & $0.80{\pm}0.55\%$ & $99.26{\pm}0.52$ & $0.90{\pm}0.59\%$ & $99.17{\pm}0.54$ & $1.00{\pm}0.62\%$ & $99.08{\pm}0.57$ \\
    RL with pretrained actor & $70.40{\pm}2.83\%$ & $42.43{\pm}2.34$ & $54.60{\pm}3.09\%$ & $54.33{\pm}2.59$ & $60.70{\pm}3.03\%$ & $50.97{\pm}2.47$ & {\rd $73.00{\pm}2.75\%$} & {\rd $43.13{\pm}2.16$} & $44.10{\pm}3.08\%$ & $63.77{\pm}2.55$ & $40.00{\pm}3.04\%$ & $69.76{\pm}2.33$ \\
    Action-space residual & {\rd $73.20{\pm}2.75\%$} & $42.98{\pm}2.17$ & $52.20{\pm}3.10\%$ & $57.37{\pm}2.54$ & $60.40{\pm}3.03\%$ & $54.03{\pm}2.34$ & {\fs\bfseries\boldmath $82.40{\pm}2.36\%$} & {\fs\bfseries\boldmath $36.41{\pm}1.85$} & $47.80{\pm}3.10\%$ & $64.67{\pm}2.33$ & $8.80{\pm}1.76\%$ & $93.54{\pm}1.30$ \\
    Single-scale prior & {\nd\slshape $73.70{\pm}2.73\%$} & {\rd $41.10{\pm}2.21$} & {\rd $59.50{\pm}3.04\%$} & {\rd $50.02{\pm}2.57$} & {\rd $65.20{\pm}2.95\%$} & {\rd $48.01{\pm}2.38$} & $72.20{\pm}2.78\%$ & $44.16{\pm}2.17$ & {\rd $67.20{\pm}2.91\%$} & {\rd $46.45{\pm}2.35$} & {\rd $65.80{\pm}2.94\%$} & {\rd $44.27{\pm}2.51$} \\
    \method frozen-decoder ablation & $72.80{\pm}2.76\%$ & {\nd\slshape $40.65{\pm}2.27$} & {\nd\slshape $61.40{\pm}3.02\%$} & {\nd\slshape $48.64{\pm}2.54$} & {\nd\slshape $67.40{\pm}2.91\%$} & {\nd\slshape $46.57{\pm}2.33$} & $72.70{\pm}2.76\%$ & $43.99{\pm}2.15$ & {\nd\slshape $72.00{\pm}2.78\%$} & {\nd\slshape $43.75{\pm}2.20$} & {\fs\bfseries\boldmath $73.30{\pm}2.74\%$} & {\nd\slshape $41.69{\pm}2.20$} \\
    \method (main) & {\fs\bfseries\boldmath $78.60{\pm}2.54\%$} & {\fs\bfseries\boldmath $35.76{\pm}2.11$} & {\fs\bfseries\boldmath $68.10{\pm}2.89\%$} & {\fs\bfseries\boldmath $42.47{\pm}2.45$} & {\fs\bfseries\boldmath $70.10{\pm}2.84\%$} & {\fs\bfseries\boldmath $43.77{\pm}2.30$} & {\nd\slshape $75.20{\pm}2.68\%$} & {\nd\slshape $41.32{\pm}2.11$} & {\fs\bfseries\boldmath $73.80{\pm}2.73\%$} & {\fs\bfseries\boldmath $43.19{\pm}2.14$} & {\nd\slshape $70.00{\pm}2.84\%$} & {\fs\bfseries\boldmath $39.23{\pm}2.48$} \\
    
    \bottomrule
    \end{tabular}}
    \end{table*}

\begin{figure*}[t]
    \centering
    \captionsetup[subfloat]{font=scriptsize}
    \setlength{\tabcolsep}{1pt}
    \begin{tabular}{@{}cccccc@{}}
    \subfloat[]{\includegraphics[width=0.155\textwidth]{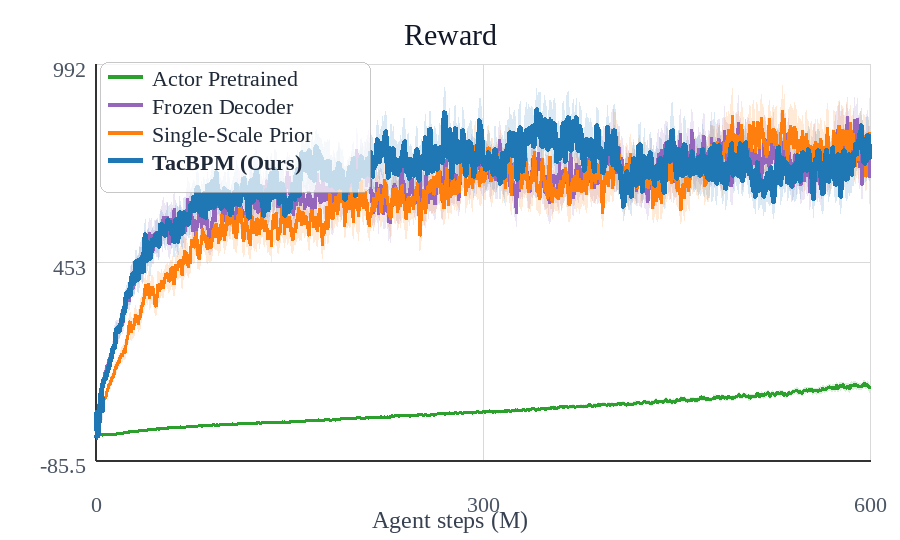}} &
    \subfloat[]{\includegraphics[width=0.155\textwidth]{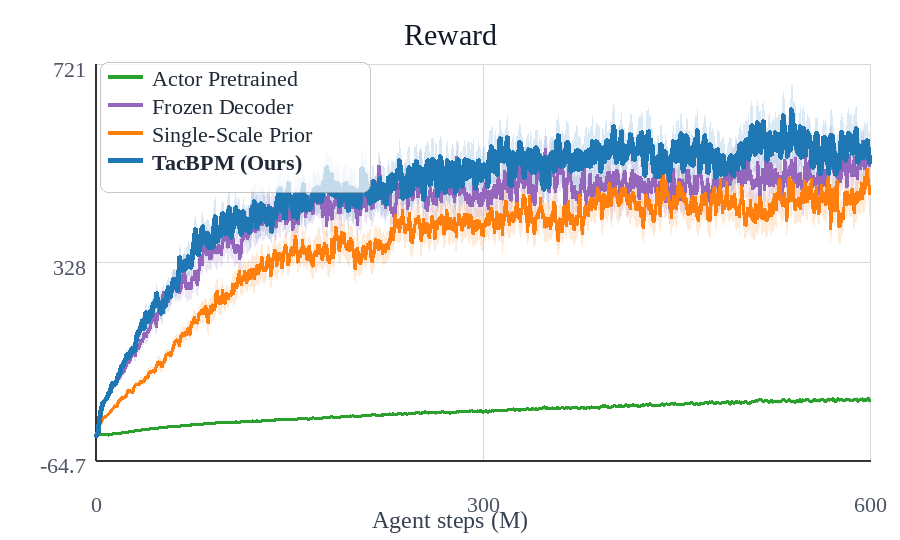}} &
    \subfloat[]{\includegraphics[width=0.155\textwidth]{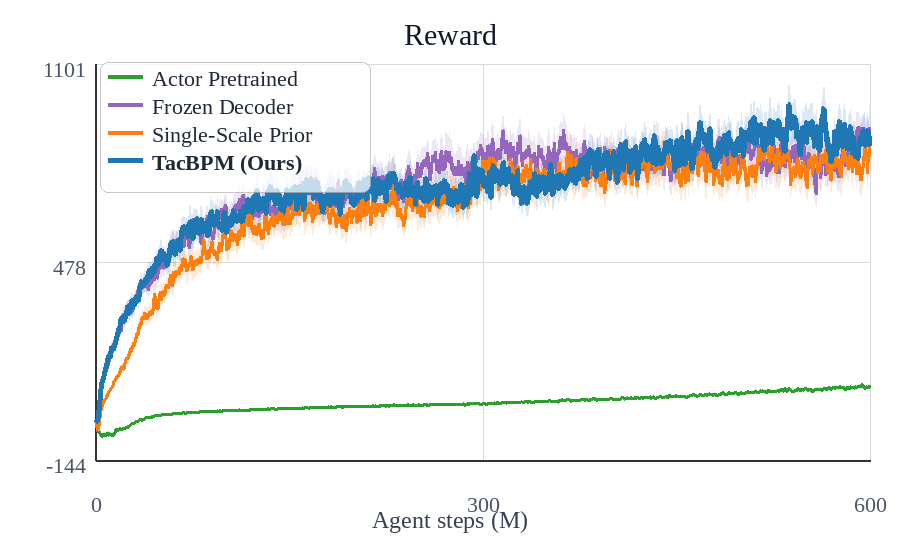}} &
    \subfloat[]{\includegraphics[width=0.155\textwidth]{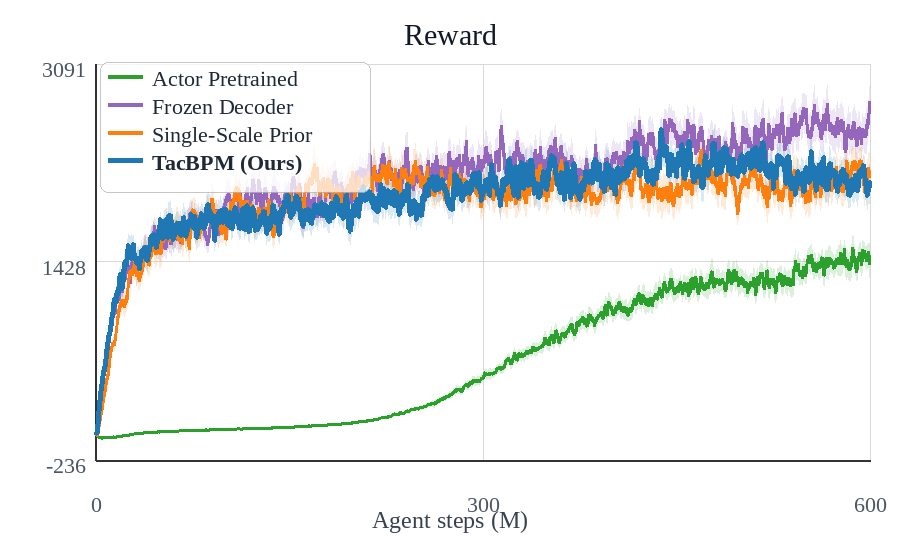}} &
    \subfloat[]{\includegraphics[width=0.155\textwidth]{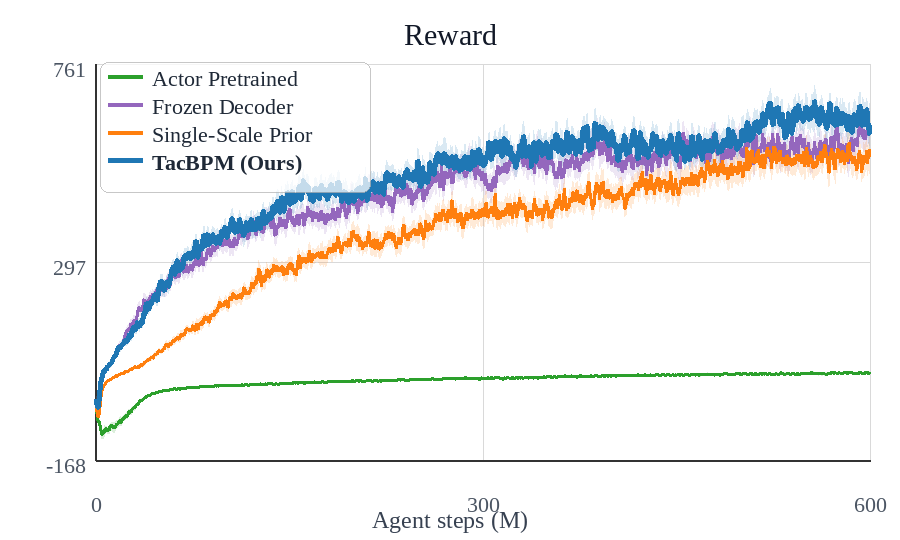}} &
    \subfloat[]{\includegraphics[width=0.155\textwidth]{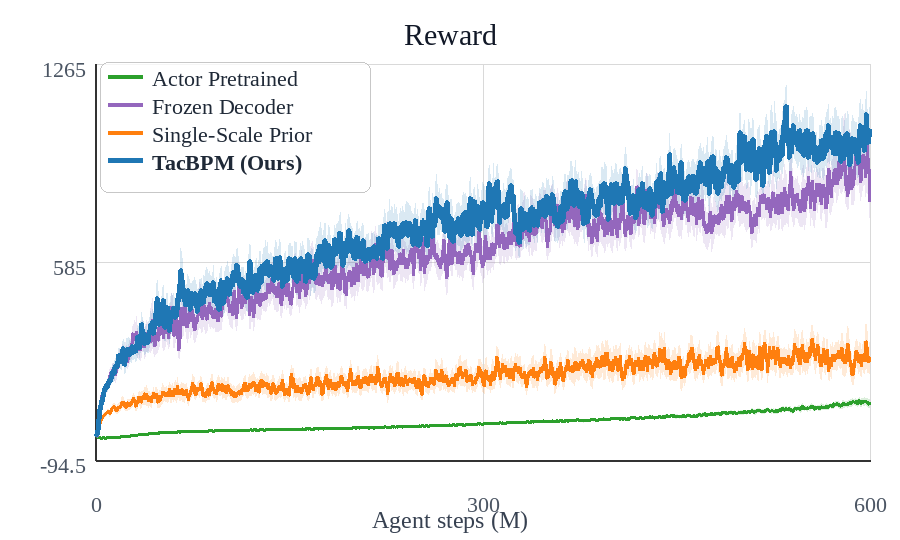}} \\
    \subfloat[]{\includegraphics[width=0.155\textwidth]{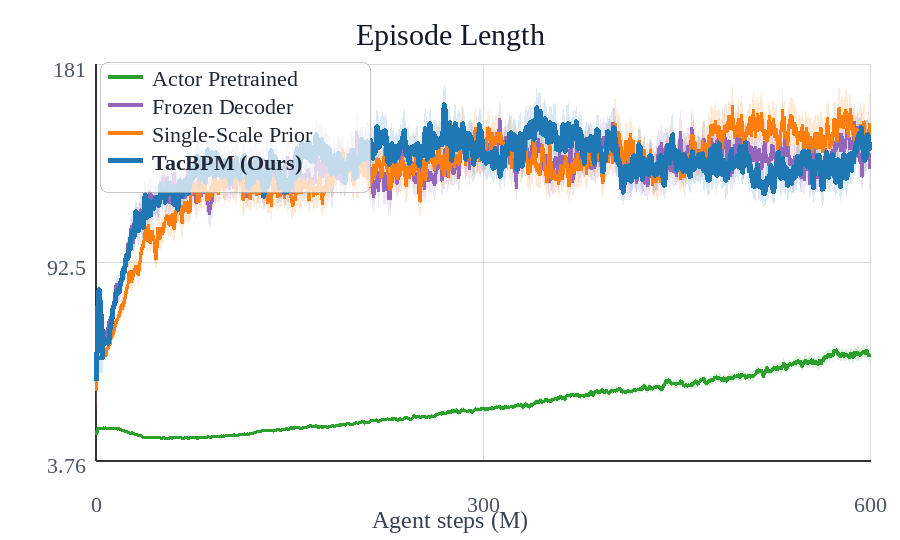}} &
    \subfloat[]{\includegraphics[width=0.155\textwidth]{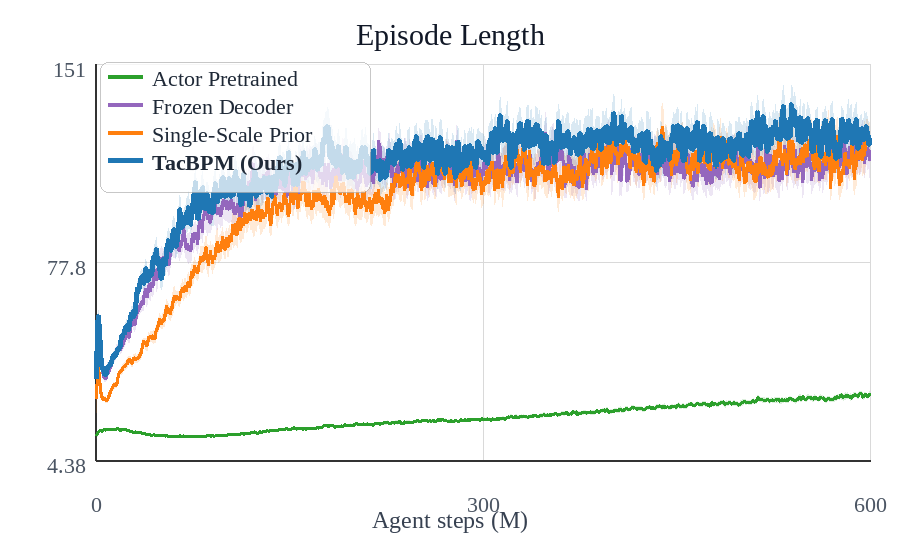}} &
    \subfloat[]{\includegraphics[width=0.155\textwidth]{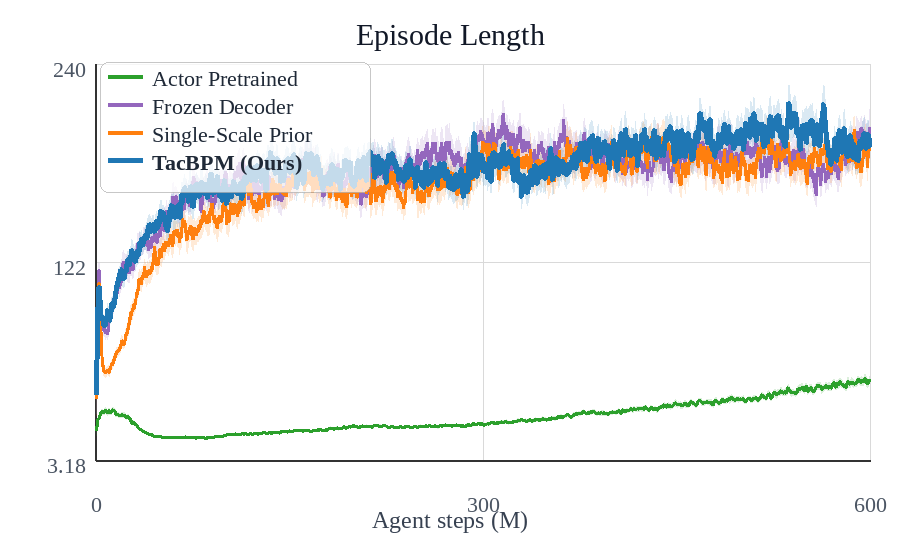}} &
    \subfloat[]{\includegraphics[width=0.155\textwidth]{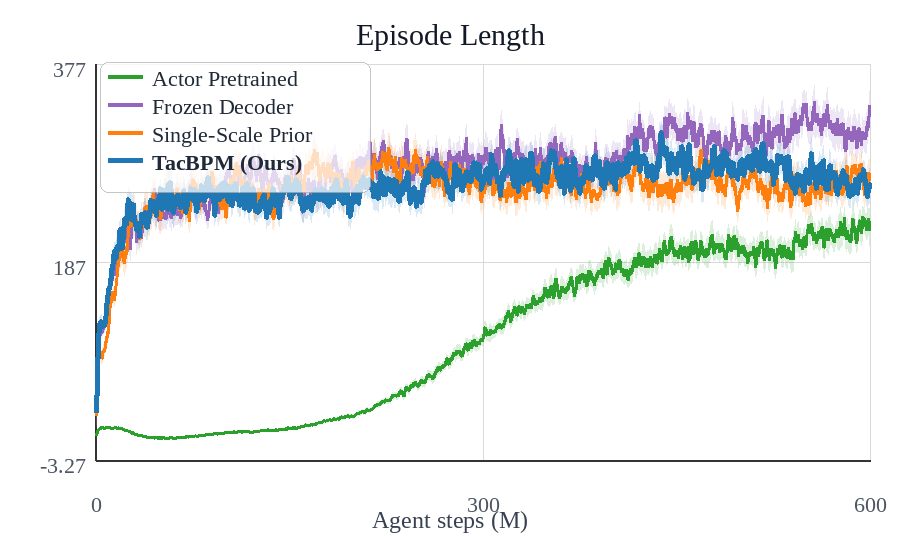}} &
    \subfloat[]{\includegraphics[width=0.155\textwidth]{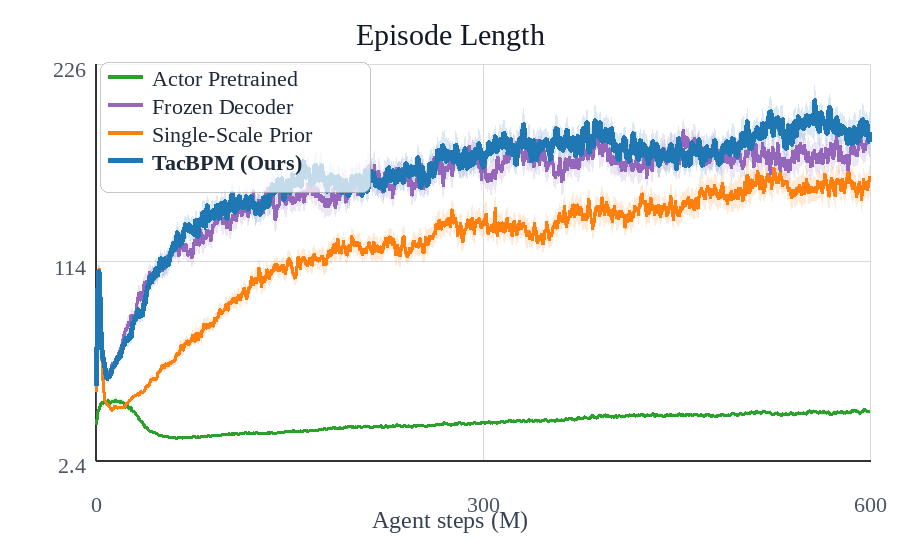}} &
    \subfloat[]{\includegraphics[width=0.155\textwidth]{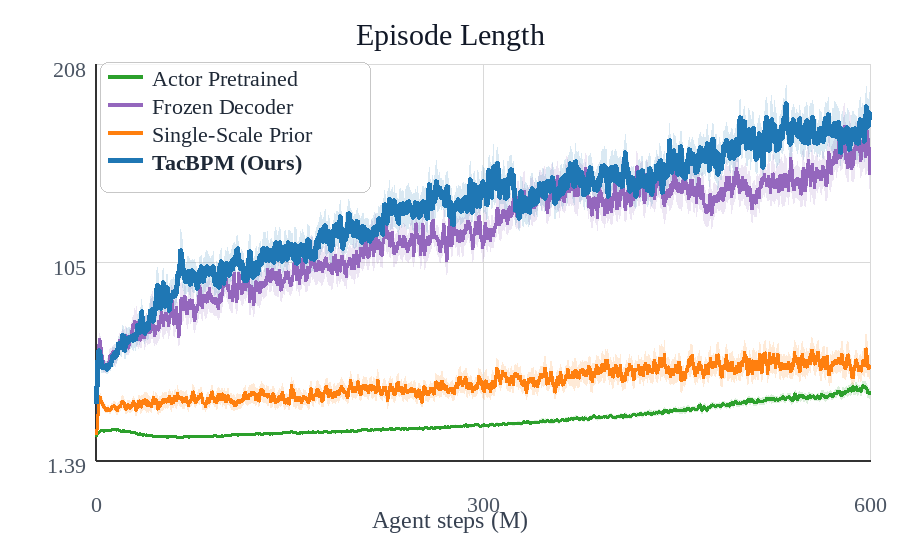}}
    \end{tabular}
    \caption{\textbf{Downstream reorientation training curves for the complex-object transfer ablations.} Columns correspond to the evaluated objects and Multi; rows show episode reward and episode length.}
    \label{fig:repose_ablation_curves}
\end{figure*}

\subsubsection{Transfer results}
Figure~\ref{fig:downstream_objects_qualitative} presents representative geometries and rollouts, with orthographic insets highlighting variation in size, aspect ratio, curvature, and local faceting. Table~\ref{tab:downstream_objects} and Fig.~\ref{fig:repose_ablation_curves} then separate action-space structure, prior diversity, decoder adaptation, and learning dynamics. The Multi setting jointly evaluates one shared policy on the five objects overlaid in Fig.~\ref{fig:prior_coverage_overview}(b), whose different scales, curvatures, elongated bodies, and faceted surfaces make contact modes substantially more heterogeneous than in single-object evaluation. Raw-action PPO rarely succeeds ($0.80$--$2.10\%$ on single objects and $1.00\%$ in Multi), with episode lengths near the failure cap, indicating that low-level exploration does not produce reliable complex-object reorientation policies within the matched budget. The gap to \method indicates that the tactile prior improves final performance, keeps early exploration near contact-stable states, and makes these matched from-scratch settings tractable downstream tasks. Actor-level transfer is stronger but transfers an executable specialist actor; it struggles in the Multi setting, where success decreases to $40.00\%$, indicating limited robustness when a single policy must accommodate several contact geometries. Action-space residual learning reaches $82.40\%$ on fruit\_02 but drops to $8.80\%$ in Multi, suggesting that low-level residuals fit favorable single geometries without preserving contact strategies across object changes. The single-scale prior is competitive, but its Multi success ($65.80\%$) remains below both multi-scale \method variants, supporting scale-diverse teacher coverage.

Decoder finetuning gives the highest single-object success on block\_03, block\_05, bottle\_01, and trash\_can\_02, improving over the frozen decoder by $5.8$, $6.7$, $2.7$, and $1.8$ percentage points. The fixed prior can therefore be retained while the decoder adapts latent commands to local geometry. The frozen decoder achieves the highest Multi success ($73.30\%$), suggesting that strict decoder reuse can regularize a shared policy over broader object variation. The learning curves further indicate that \method variants improve while maintaining longer controlled interactions.

\vspace{-2mm}
\subsection{Axis-Conditioned Rotation}
\vspace{-2mm}

\subsubsection{Commanded-axis protocol}
Axis-Conditioned Rotation tests whether the same latent controller can follow compact task commands. A single policy rotates the object about one of six signed Cartesian axes ($+x$, $-x$, $+y$, $-y$, $+z$, and $-z$) while maintaining stable contact and motion gaits. The reward encourages commanded-axis angular velocity and penalizes orthogonal rotation, so the policy must learn direction-specific rolling rather than arbitrary object motion. The policy freezes the VAE prior and finetunes the decoder under object, contact, mass, actuation, noise, push, and gravity randomization. Each command is evaluated independently; logs include signed, total, perpendicular rotation and drop rate, while Table~\ref{tab:axis_conditioned_rotation} reports the macro-average signed angle.

\subsubsection{Simulation results}
\begin{figure}[t]
    \centering
    \setlength{\tabcolsep}{1.2pt}
    \begin{tabular}{@{}ccc@{}}
    \includegraphics[width=0.318\columnwidth]{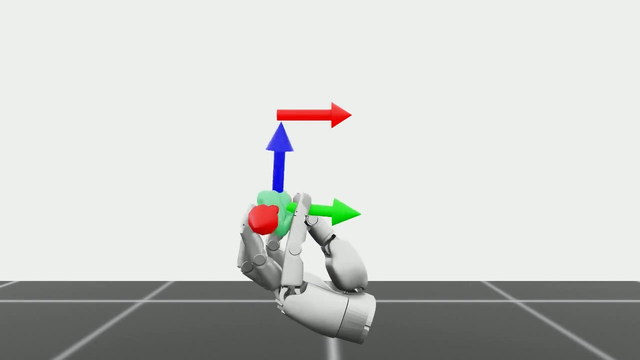} &
    \includegraphics[width=0.318\columnwidth]{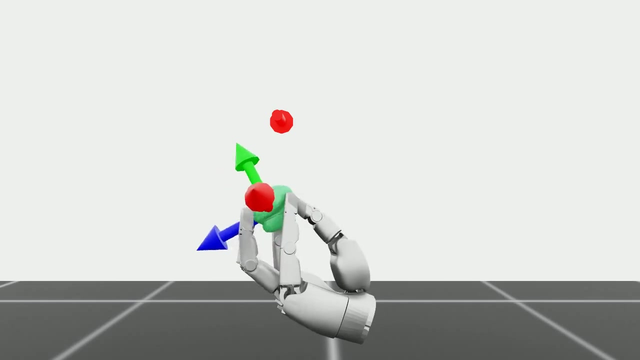} &
    \includegraphics[width=0.318\columnwidth]{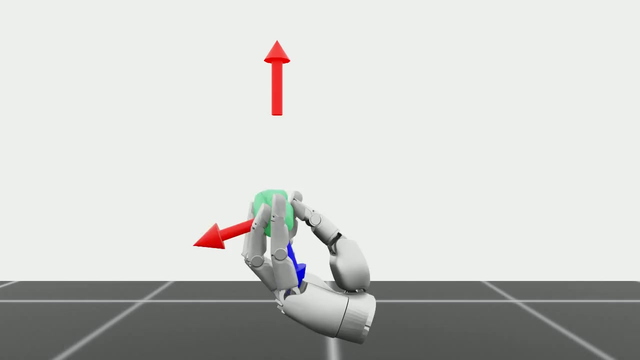} \\
    {\scriptsize (a) $+x$} & {\scriptsize (b) $+y$} & {\scriptsize (c) $+z$} \\
    \includegraphics[width=0.318\columnwidth]{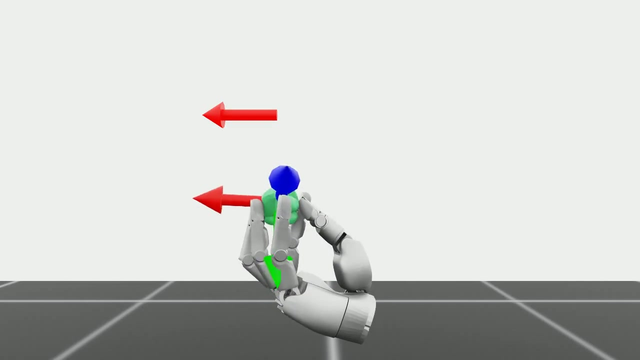} &
    \includegraphics[width=0.318\columnwidth]{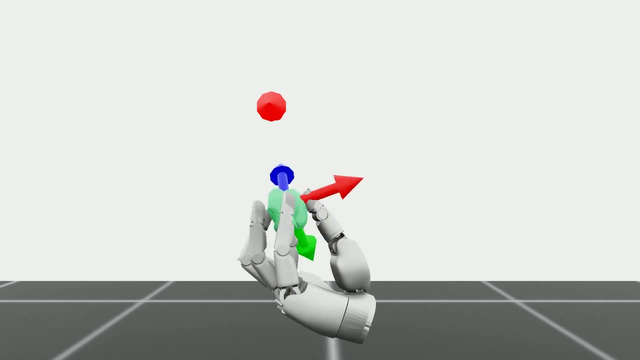} &
    \includegraphics[width=0.318\columnwidth]{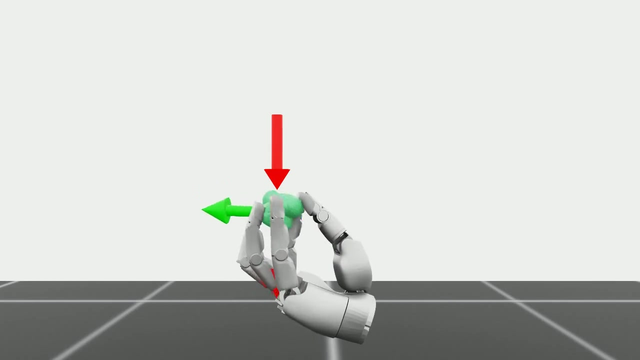} \\
    {\scriptsize (d) $-x$} & {\scriptsize (e) $-y$} & {\scriptsize (f) $-z$}
    \end{tabular}
    \caption{\textbf{Commanded-axis simulation examples.} The panels show rollouts for the six signed Cartesian rotation commands.}
    \label{fig:axis_condition_qualitative}
    \vspace{-2mm}

\end{figure}

Figure~\ref{fig:axis_condition_qualitative} illustrates commanded rollouts for all six signed axes, showing that the same policy must change both direction and contact strategy without switching controllers.

\begin{table}[t]
\caption{\textbf{Commanded-axis rotation on six simulated objects.} Axis Avg. is the macro-average signed commanded-axis angle over the six commands.}
\label{tab:axis_conditioned_rotation}
\centering
\setlength{\tabcolsep}{2.2pt}
\renewcommand{\arraystretch}{1.05}
\resizebox{\columnwidth}{!}{%
\begin{tabular}{lccc}
\toprule
Object & From scratch & No tactile & Ours \\
\midrule
trashcan\_02 & $76.63{\pm}42.48$ & $48.74{\pm}16.78$ & {\fs\bfseries\boldmath $225.61{\pm}169.78$} \\
strawberry & $59.44{\pm}14.64$ & $44.10{\pm}13.64$ & {\fs\bfseries\boldmath $69.39{\pm}10.56$} \\
bottle\_02 & $-0.43{\pm}14.65$ & $181.68{\pm}56.49$ & {\fs\bfseries\boldmath $192.98{\pm}43.85$} \\
block\_03 & $64.34{\pm}12.42$ & $75.24{\pm}18.87$ & {\fs\bfseries\boldmath $78.28{\pm}25.67$} \\
block\_08 & $1.29{\pm}18.48$ & $80.21{\pm}11.50$ & {\fs\bfseries\boldmath $117.39{\pm}27.80$} \\
block\_14 & $0.00{\pm}9.94$ & $49.55{\pm}21.90$ & {\fs\bfseries\boldmath $77.06{\pm}31.96$} \\
\bottomrule
\end{tabular}}
\end{table}

Across the six simulated objects, \method obtains the highest axis-average rotation. Raw-action exploration struggles on bottle\_02, block\_08, and block\_14, where the signed angle is near zero, indicating that the policy often fails to discover a commanded rolling mode before contact degrades. The no-tactile variant often exceeds raw-action exploration, indicating that the prior-decoder framework contributes structure even without tactile input; the remaining gap to \method reflects the value of contact-state information under geometry and support-mode variation. The large variability reflects heterogeneous off-axis or contact-loss outcomes.

\subsubsection{Real-robot protocol and results}
\begin{figure}[t]
    \centering
    \setlength{\tabcolsep}{1.2pt}
    \begin{tabular}{@{}ccc@{}}
    \includegraphics[width=0.305\columnwidth]{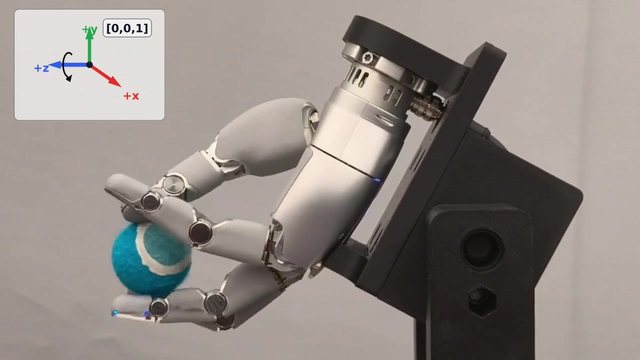} &
    \includegraphics[width=0.305\columnwidth]{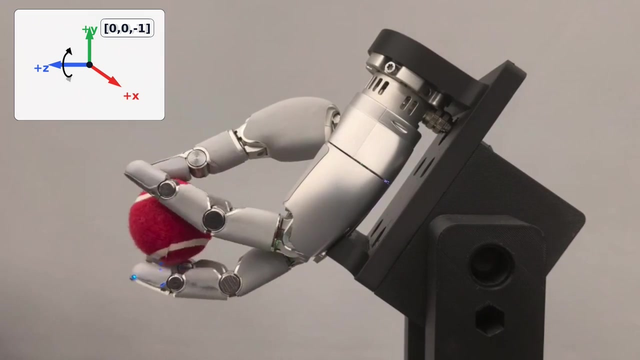} &
    \includegraphics[width=0.305\columnwidth]{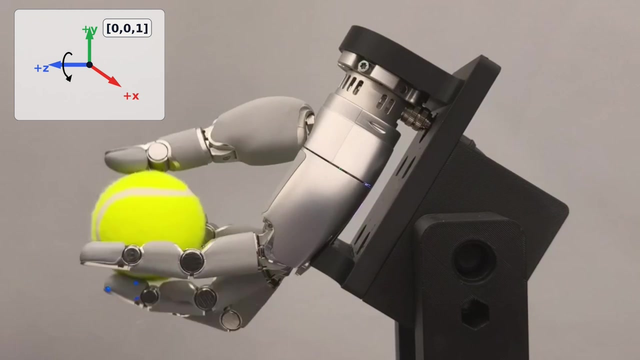} \\
    {\scriptsize (a) small tennis} & {\scriptsize (b) middle tennis} & {\scriptsize (c) standard tennis} \\
    \includegraphics[width=0.305\columnwidth]{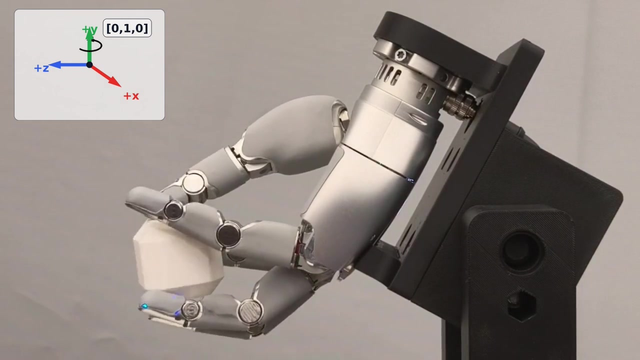} &
    \includegraphics[width=0.305\columnwidth]{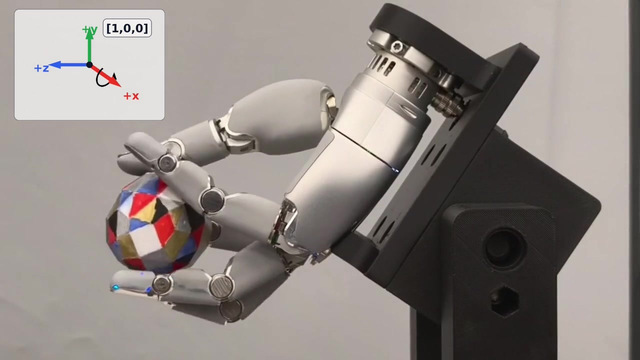} &
    \includegraphics[width=0.305\columnwidth]{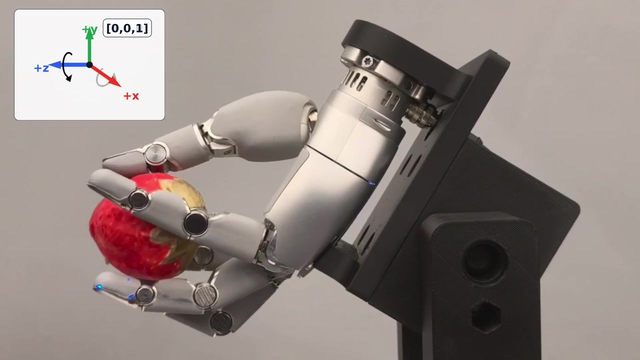} \\
    {\scriptsize (d) corner block} & {\scriptsize (e) multiface} & {\scriptsize (f) strawberry}
    \end{tabular}
    \caption{\textbf{Real-robot commanded-axis rollouts.} The panels show physical objects used in the hardware evaluation.}
    \label{fig:real_axis_qualitative}
    \vspace{-2mm}

\end{figure}

\begin{table}[t]
\caption{\textbf{Real-robot commanded-axis rotation.} Entries report mean signed rotation and success for each commanded direction.}
\label{tab:real_axis_conditioned_rotation}
\centering
\setlength{\tabcolsep}{1.9pt}
\renewcommand{\arraystretch}{1.05}
{\scriptsize
\newcommand{\axisbest}[1]{\begingroup\setlength{\fboxsep}{0.6pt}\colorbox{colorFst}{\strut #1}\endgroup}
\resizebox{\columnwidth}{!}{%
\begin{tabular}{llcccccc}
\toprule
Object & Method & \makecell{$+x$\\Rot., Succ.} & \makecell{$-x$\\Rot., Succ.} & \makecell{$+y$\\Rot., Succ.} & \makecell{$-y$\\Rot., Succ.} & \makecell{$+z$\\Rot., Succ.} & \makecell{$-z$\\Rot., Succ.} \\
\midrule
\multirow{3}{*}{corner block}
& From scratch & $428^\circ$, 6/10 & $162^\circ$, 3/10 & $27^\circ$, 0/10 & $81^\circ$, 1/10 & $293^\circ$, 4/10 & $113^\circ$, 2/10 \\
& No tactile & $59^\circ$, 1/10 & $22^\circ$, 0/10 & $31^\circ$, 0/10 & $13^\circ$, 0/10 & $257^\circ$, 4/10 & $30^\circ$, 0/10 \\
& Ours & \axisbest{\bfseries\boldmath $644^\circ$}, \axisbest{\bfseries 9/10} & \axisbest{\bfseries\boldmath $524^\circ$}, \axisbest{\bfseries 8/10} & \axisbest{\bfseries\boldmath $753^\circ$}, \axisbest{\bfseries 10/10} & \axisbest{\bfseries\boldmath $441^\circ$}, \axisbest{\bfseries 7/10} & \axisbest{\bfseries\boldmath $806^\circ$}, \axisbest{\bfseries 10/10} & \axisbest{\bfseries\boldmath $405^\circ$}, \axisbest{\bfseries 7/10} \\
\midrule
\multirow{3}{*}{small tennis}
& From scratch & $243^\circ$, 4/10 & $171^\circ$, 3/10 & $54^\circ$, 1/10 & $149^\circ$, 2/10 & $441^\circ$, 7/10 & $342^\circ$, 5/10 \\
& No tactile & $337^\circ$, 5/10 & \axisbest{\bfseries\boldmath $528^\circ$}, \axisbest{\bfseries 8/10} & $19^\circ$, 0/10 & $14^\circ$, 0/10 & $576^\circ$, 8/10 & \axisbest{\bfseries\boldmath $473^\circ$}, \axisbest{\bfseries 7/10} \\
& Ours & \axisbest{\bfseries\boldmath $765^\circ$}, \axisbest{\bfseries 10/10} & $519^\circ$, \axisbest{\bfseries 8/10} & \axisbest{\bfseries\boldmath $383^\circ$}, \axisbest{\bfseries 6/10} & \axisbest{\bfseries\boldmath $338^\circ$}, \axisbest{\bfseries 5/10} & \axisbest{\bfseries\boldmath $945^\circ$}, \axisbest{\bfseries 10/10} & $394^\circ$, 6/10 \\
\midrule
\multirow{3}{*}{standard tennis}
& From scratch & $70^\circ$, 1/10 & $40^\circ$, 0/10 & $288^\circ$, 4/10 & $38^\circ$, 0/10 & $216^\circ$, 3/10 & $342^\circ$, 5/10 \\
& No tactile & $59^\circ$, 1/10 & $41^\circ$, 0/10 & $202^\circ$, 3/10 & $132^\circ$, \axisbest{\bfseries 2/10} & $218^\circ$, 3/10 & $239^\circ$, 4/10 \\
& Ours & \axisbest{\bfseries\boldmath $378^\circ$}, \axisbest{\bfseries 6/10} & \axisbest{\bfseries\boldmath $192^\circ$}, \axisbest{\bfseries 3/10} & \axisbest{\bfseries\boldmath $495^\circ$}, \axisbest{\bfseries 7/10} & \axisbest{\bfseries\boldmath $137^\circ$}, \axisbest{\bfseries 2/10} & \axisbest{\bfseries\boldmath $456^\circ$}, \axisbest{\bfseries 7/10} & \axisbest{\bfseries\boldmath $633^\circ$}, \axisbest{\bfseries 9/10} \\
\midrule
\multirow{3}{*}{\makecell{multiface object\\(unseen)}}
& From scratch & $158^\circ$, 2/10 & $22^\circ$, 0/10 & $315^\circ$, 5/10 & $279^\circ$, 4/10 & $167^\circ$, 3/10 & $16^\circ$, 0/10 \\
& No tactile & $258^\circ$, 4/10 & $297^\circ$, 4/10 & $23^\circ$, 0/10 & $48^\circ$, 0/10 & $162^\circ$, 3/10 & $239^\circ$, 4/10 \\
& Ours & \axisbest{\bfseries\boldmath $675^\circ$}, \axisbest{\bfseries 9/10} & \axisbest{\bfseries\boldmath $459^\circ$}, \axisbest{\bfseries 7/10} & \axisbest{\bfseries\boldmath $797^\circ$}, \axisbest{\bfseries 10/10} & \axisbest{\bfseries\boldmath $324^\circ$}, \axisbest{\bfseries 5/10} & \axisbest{\bfseries\boldmath $754^\circ$}, \axisbest{\bfseries 10/10} & \axisbest{\bfseries\boldmath $799^\circ$}, \axisbest{\bfseries 10/10} \\
\bottomrule
\end{tabular}}}
\end{table}

The real-robot results in Fig.~\ref{fig:real_axis_qualitative} and Table~\ref{tab:real_axis_conditioned_rotation} exhibit a similar trend under hardware contact noise. Deployment uses the simulation observation structure with calibrated tactile forces and contact points, runs proprioception and tactile feedback at \SI{20}{Hz}, and matches joint gains to measured \SI{1}{Hz} sinusoidal responses; object-pose feedback is not used in the in-hand action loop. An episode succeeds if the object rotates more than $180^\circ$ along the commanded direction within \SI{20}{s}; drops or stuck states before the threshold are considered as failures. Each object-axis entry uses 10 episodes. \method achieves the strongest performance on most object-axis pairs, with pronounced improvements on corner block, standard tennis, and the unseen multiface object. The no-tactile policy struggles on the $y$-axis commands for corner block and small tennis, consistent with a loss of directional contact information during lateral rolling. The small-tennis exceptions favor no-tactile in accumulated angle, but rollouts show less stable contact, including slips, drops, and stuck configurations. Because accumulated angle does not fully measure stability, the videos complement Table~\ref{tab:real_axis_conditioned_rotation} by showing more consistent rolling contact\footnote{Project videos: \projectwebsite.}.

\noindent\textbf{Qualitative failure modes.}
The real-robot failures mainly involve contact drift, slow off-axis motion, out-of-distribution hand-object configurations, drops, or objects wedged between fingers; tactile conditioning reduces these cases by grounding latent commands in current contact. Command-switch transients remain challenging: in a corner-block rollout, switching from $-x$ to $+z$ before stable contact is re-established causes a drop, suggesting the need for transition handling or stability-aware command gating.

\vspace{-2mm}
\subsection{Grasp-to-AnyPose Arm-Hand Evaluation}
\vspace{-2mm}

\subsubsection{Task and evaluation}
Following the Grasp-to-AnyPose setup in Sec.~\ref{sec:arm_hand_method}, 
we evaluate object-category robustness from the rubber-hammer teacher family to small hammer, blue brush, staples marker, and mallet hammer. 
Single-object evaluations use 1024 deterministic episodes, and count failed episodes as \SI{20}{cm} position error and \SI{20}{\degree} rotation error.

\begin{figure}[t]
    \centering
    \includegraphics[width=0.93\columnwidth]{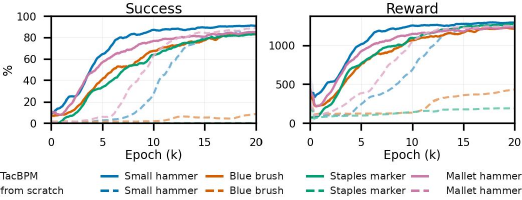}
    \caption{\textbf{Arm-hand Grasp-to-AnyPose training curves.} The panels show success rate and episode reward across four novel tool geometries.}
    \label{fig:arm_hand_learning_curves}
    \vspace{-2mm}
\end{figure}

Figure~\ref{fig:arm_hand_learning_curves} compares per-object learning dynamics. Under the same reward, curriculum, and PPO settings, the residual-latent policy reaches higher success and reward earlier than raw-action exploration, including on the more challenging blue-brush and staples-marker geometries.
These cases are diagnostic because raw-action exploration often fails to discover stable grasp-to-goal behavior within the same training horizon.

\subsubsection{Simulation evaluation and hardware demonstration}
The quantitative arm-hand results are evaluated in simulation on novel tool geometries. We also deploy the same policy on hardware without additional real-robot policy finetuning after the simulation-to-real calibration used for the hand experiments, using FoundationPose object pose and forward-kinematic end-effector pose to construct the object-relative keypoints and goal features. This qualitative hardware result demonstrates successful sim-to-real transfer for the complete grasp--lift--goal sequence rather than a large-scale real-robot benchmark.

\begin{figure}[!t]
    \centering
    \includegraphics[width=0.88\columnwidth]{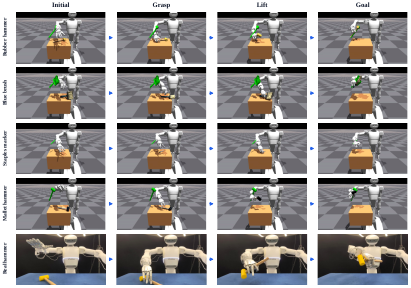}
    \caption{\textbf{Arm-hand Grasp-to-AnyPose qualitative evaluation.} Columns show the Initial--Grasp--Lift--Goal progression for simulated and real-robot rollouts.}
    \label{fig:arm_hand_first_goal_sequence}
    \vspace{-2mm}
\end{figure}

\begin{table}[!t]
\caption{\textbf{Arm-hand simulation evaluation on diverse tools.} Metrics are success (\%), position error (cm), and rotation error (deg).}
\label{tab:arm_hand}
\centering
{\setlength{\tabcolsep}{2.6pt}
\renewcommand{\arraystretch}{1.05}
\resizebox{\columnwidth}{!}{%
\begin{tabular}{lcccccc}
\toprule
\multirow{2}{*}{Object} & \multicolumn{3}{c}{RL from scratch} & \multicolumn{3}{c}{\method} \\
\cmidrule(lr){2-4}\cmidrule(lr){5-7}
& Succ. $\uparrow$ & Pos. err. $\downarrow$ & Rot. err. $\downarrow$ & Succ. $\uparrow$ & Pos. err. $\downarrow$ & Rot. err. $\downarrow$ \\
\midrule
Small hammer & $77.73{\pm}2.55$ & $8.73{\pm}0.46$ & $11.09{\pm}0.49$ & {\fs\bfseries\boldmath $97.56{\pm}0.95$} & {\fs\bfseries\boldmath $5.46{\pm}0.23$} & {\fs\bfseries\boldmath $8.48{\pm}0.48$} \\
Blue brush & $27.44{\pm}2.73$ & $16.29{\pm}0.41$ & $18.09{\pm}0.42$ & {\fs\bfseries\boldmath $95.70{\pm}1.24$} & {\fs\bfseries\boldmath $5.55{\pm}0.29$} & {\fs\bfseries\boldmath $11.82{\pm}0.76$} \\
Staples marker & $0.49{\pm}0.43$ & $19.96{\pm}0.04$ & $19.95{\pm}0.05$ & {\fs\bfseries\boldmath $97.36{\pm}0.98$} & {\fs\bfseries\boldmath $4.71{\pm}0.20$} & {\fs\bfseries\boldmath $7.30{\pm}0.40$} \\
Mallet hammer & $91.31{\pm}1.73$ & $5.93{\pm}0.34$ & $11.30{\pm}0.61$ & {\fs\bfseries\boldmath $96.29{\pm}1.16$} & {\fs\bfseries\boldmath $5.55{\pm}0.29$} & {\fs\bfseries\boldmath $10.43{\pm}0.69$} \\
\bottomrule
\end{tabular}}}
\end{table}

Figure~\ref{fig:arm_hand_first_goal_sequence} shows the Initial--Grasp--Lift--Goal progression for the rubber-hammer teacher family, held-out simulated tools, and a real-robot hammer rollout; videos show additional generalized initial placements.
Table~\ref{tab:arm_hand} quantifies the simulated evaluation on diverse tools. \method improves success on all four categories and substantially reduces capped position and rotation errors on blue brush and staples marker, where matched raw-action PPO rarely reaches stable goal-pose behavior. The staples marker is the most pronounced failure case for raw-action learning, with $0.49\%$ success and errors saturated near the cap, indicating that the baseline seldom completes a stable grasp before goal reaching. These gains indicate that the proposed framework accelerates optimization and raises the empirical performance ceiling under the same protocol, while Fig.~\ref{fig:arm_hand_first_goal_sequence} and the project videos provide complementary evidence of successful sim-to-real transfer.

\FloatBarrier
\vspace{-3mm}
\section{Conclusion}
\vspace{-0.5mm}

We presented \method, a tactile-conditioned behavior prior model for dexterous reorientation. 
By distilling multi-scale sphere specialists into a tactile-proprioceptive latent controller, \method gives downstream policies a compact residual control framework anchored near contact-stable behavior. 
Across in-hand transfer, commanded-axis rotation, and a separate arm-hand extension, the results indicate improved exploration over raw joint-action learning, stronger adaptation under contact and geometry shifts, and effective reuse of the teacher-distillation and residual-latent paradigm.
Decoder adaptation further supports in-hand transfer to new geometries while preserving the contact-conditioned prior.
The arm-hand study further suggests that the paradigm can be applied across embodiments with separately trained teachers and priors.
Overall, simulation and real-robot results indicate that tactile-conditioned behavior priors are effective for reorientation-style manipulation.

\vspace{-3mm}

\bibliographystyle{IEEEtrans}
\bibliography{root}

\end{document}